\documentclass{article}

\PassOptionsToPackage{numbers,sort&compress}{natbib}
\PassOptionsToPackage{table}{xcolor}
\usepackage[preprint]{iron_report}

\usepackage[utf8]{inputenc}
\usepackage[T1]{fontenc}
\usepackage{amsmath}
\usepackage{amsfonts}
\usepackage{amssymb}
\usepackage{booktabs}
\usepackage{fancyhdr}
\usepackage{graphicx}
\usepackage{microtype}
\usepackage{nicefrac}
\usepackage{tabularx}
\usepackage{xcolor}
\usepackage{hyperref}
\usepackage{url}
\usepackage{tcolorbox}

\usepackage{multirow}
\usepackage{makecell}
\usepackage{multicol}
\usepackage{tablefootnote}

\definecolor{xpgreen}{HTML}{96B414}
\definecolor{xpgdarkgreen}{HTML}{234100}
\definecolor{xpgpalegreen}{HTML}{E6F0DC}
\definecolor{xpgblack}{HTML}{000000}
\definecolor{xpgdarkgray}{HTML}{4B4B4B}
\definecolor{xpglightgray}{HTML}{E1E1E1}

\colorlet{ironblue}{xpgdarkgreen}
\colorlet{ironcyan}{xpgdarkgreen}
\colorlet{ironink}{xpgblack}
\colorlet{ironmuted}{xpgdarkgray}
\colorlet{ironlight}{xpgpalegreen}
\colorlet{ironline}{xpglightgray}

\hypersetup{
  colorlinks=true,
  linkcolor=ironblue,
  citecolor=ironblue,
  urlcolor=ironblue
}

\newcommand{\modelname}{IronLLM}

\renewcommand{\arraystretch}{1.12}

\newcommand{\reportlogo}{\colorbox{white}{\includegraphics[height=0.8em]{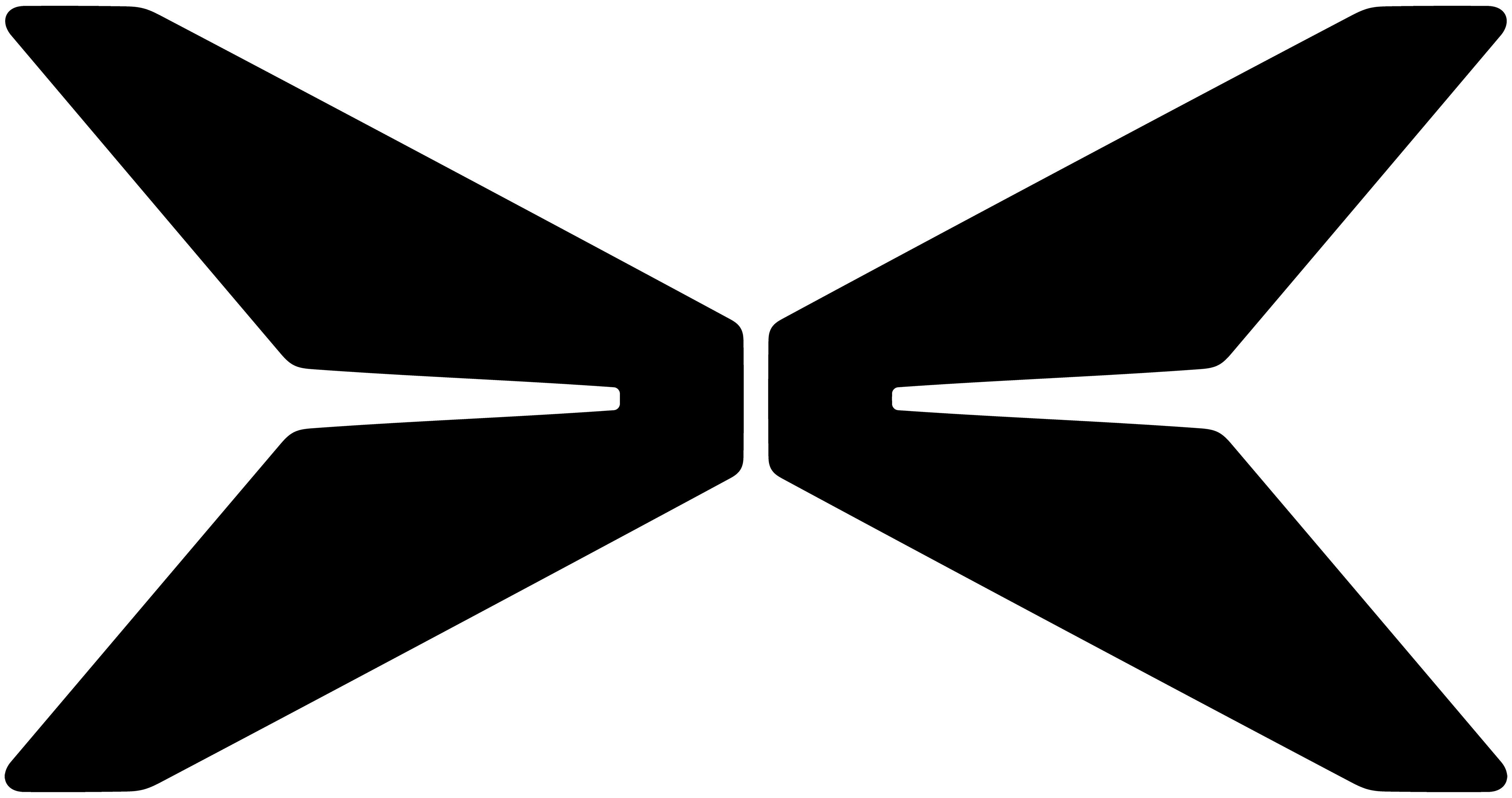}}}
\newcommand{\xpengbrandlockup}{%
  \includegraphics[height=5.2mm]{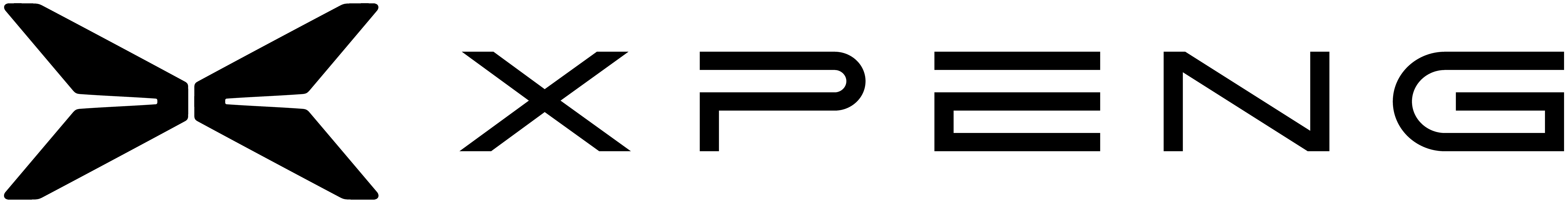}%
}

\renewcommand{\headwidth}{\textwidth}

\renewcommand{\headrulewidth}{0.5pt}
\renewcommand{\headrule}{%
  \vspace{2pt}%
  \hbox to\headwidth{%
    \color{black}%
    \leaders\hrule height\headrulewidth\hfill
  }%
}

\fancypagestyle{titlepage}{%
  \fancyhf{}%
  \fancyhead[L]{%
    \raisebox{-20pt}{\xpengbrandlockup}%
  }%
  \renewcommand{\headrulewidth}{0pt}%
}

\title{\modelname{}: Forging Compact Edge-Native Language Models for Real-Time Embodied Intelligence}

\author{%
  \parbox{0.94\textwidth}{%
    \normalfont
    Changdi Yang\textsuperscript{*},
    Fengquan Jiao\textsuperscript{*},
    Haochih Lin\textsuperscript{*},
    Haoran Yang\textsuperscript{*},
    Jing Xiao\textsuperscript{*},
    Liangyu Huo\textsuperscript{*},
    Suxin Lu\textsuperscript{*},
    Tiance Chen\textsuperscript{*},
    Wei Liu\textsuperscript{*},
    Yinggan Xu\textsuperscript{*},
    Yunxiang Lu\textsuperscript{*\dag},
    Zai Zheng\textsuperscript{*},
    Zhirui Xie\textsuperscript{*},
    Zhongyang Che\textsuperscript{*\dag},
    Ziyan Tang\textsuperscript{*\dag},
    Zuoxiang Zhao\textsuperscript{*},
    Jian Yao\textsuperscript{\ddag}}\\[5mm]
    \textbf{Robotics Foundation Model Team, Xpeng Inc.}
}

\begin{document}

\maketitle
\thispagestyle{titlepage}
\enlargethispage{2\baselineskip}
\begingroup
  \renewcommand{\thefootnote}{\fnsymbol{footnote}}
  \footnotetext[1]{Equal contribution, listed alphabetically by first name.}
  \footnotetext[2]{Technical leadership.}
  \footnotetext[3]{Supervision.}
\endgroup

\vspace{-0.5em}

\begin{abstract}
    We present \textbf{IronLLM-0.6B}, a 654M-parameter language model designed for efficient on-device inference. IronLLM-0.6B combines a hybrid attention architecture with \textbf{X-MTP}, a lightweight shared-KV multi-token prediction design that eliminates per-depth KV-cache replay and employs a lightweight verification head for rollback-free drafting, achieving a 1.48x decoding speedup. The model is pretrained on approximately 6.2 trillion tokens using a quality-oriented data pipeline and is further post-trained with Multi-Domain On-Policy Distillation to integrate capabilities from domain-specialized teachers. To better meet the low-latency requirements of on-device scenarios, IronLLM-0.6B adopts an \textbf{Instruct-Only} design. Evaluations show that IronLLM-0.6B achieves competitive performance relative to larger models such as Qwen3.5-0.8B and MiniCPM5-1B, while producing more concise responses on many tasks. We further present \textbf{IronLLM-0.6B-Light}, which replaces RMSNorm with Dynamic Tanh and simplifies several computationally expensive components to improve inference and quantization efficiency. Together, the IronLLM models provide an effective performance--efficiency trade-off for resource-constrained deployment.
    
    \par\vspace{1.5ex}
        
    \noindent
    \begingroup
    \small
    \setlength{\tabcolsep}{0pt}
    \renewcommand{\arraystretch}{1.25}
    \begin{tabular}{
      @{}
      l
      @{\hspace{0.4em}}
      l
      @{\hspace{0.45em}}
      l
      @{}
    }
        \hspace{-0.5em}
        \makebox[0.9em][c]{%
          \raisebox{-0.18\height}{%
            \includegraphics[
              width=1.08em,
              height=1.08em,
              keepaspectratio
            ]{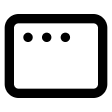}%
          }%
        }
        &
        \textbf{Project:}
        &
        \href{https://xpeng-robotics.github.io/iron-fm}
        {\texttt{https://xpeng-robotics.github.io/iron-fm}}
    \end{tabular}
    \endgroup
\end{abstract}

\begin{figure}[ht]
    \centering
    \includegraphics[width=\linewidth]{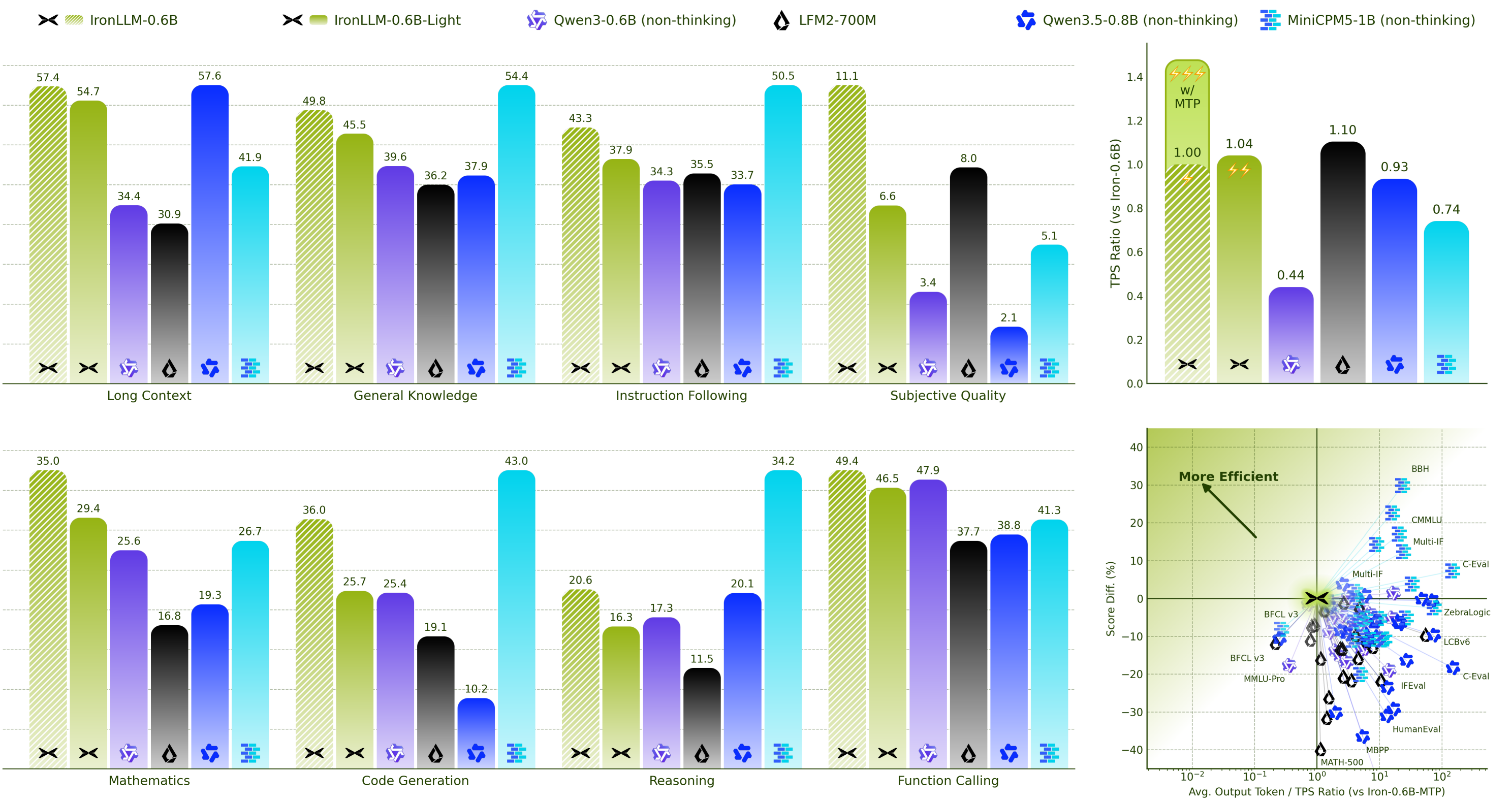}
    \vspace{-1.0em}
    \caption{\textbf{Capability and inference efficiency of compact language models.} Left: scores across eight evaluation categories. Upper right: decoding throughput relative to IronLLM-0.6B at 32K context. Lower right: benchmark-level score differences versus relative generation time, compared with IronLLM-0.6B MTP.}
    \label{fig:model_score_and_relative_efficiency}
    \vspace{-1em}
\end{figure}

\clearpage
\thispagestyle{fancy}
\tableofcontents
\clearpage

\section{Introduction}

The rapid advancement of large language models (LLMs) has led to unprecedented capabilities in natural language understanding, reasoning, and generation, with today's frontier models scaling to hundreds of billions or even trillions of parameters~\cite{deepseekai2024deepseekv3,yang2025qwen3,zeng2026glm,kimiteam2026kimik3openfrontier}. However, the practical deployment of AI assistants in real-world settings demands a fundamentally different set of priorities. This is particularly evident in embodied intelligence, where robots, in-vehicle systems, and smart cockpits must understand instructions, reason about their surroundings, and respond in real time. In these on-device contexts, factors such as inference latency, memory footprint, energy consumption, and operational reliability often outweigh the raw capacity offered by massive scale. Consequently, embodied intelligence creates a critical need for compact, high-performance models that deliver robust capabilities entirely on the edge, without reliance on network connectivity or remote computation.

Achieving strong performance within a severely constrained parameter budget poses unique challenges across the model development lifecycle. On the data front, small models exhibit heightened sensitivity to corpus quality and distribution, requiring substantially more efficient data curation than their larger counterparts, for which massive data scale can often compensate for noise~\cite{hu2024minicpmunveilingpotentialsmall}. Architecturally, components that yield marginal gains in large models may introduce disproportionate overhead in compact regimes, calling for careful co-design of model structure and inference efficiency under stringent onboard compute and power budgets. Furthermore, post-training paradigms must be tailored to preserve and integrate diverse capabilities without inducing the catastrophic forgetting or cross-domain interference that small models are particularly susceptible to. These considerations motivate treating on-device deployment for embodied applications not as an afterthought but as a first-class design constraint spanning data, architecture, and training methodology.

In this work, we introduce \textbf{IronLLM-0.6B}, a compact yet high-performance large language model with a hybrid attention architecture~\cite{qwen3.5,merrill2026olmohybridtheorypractice,kimiteam2026kimik3openfrontier}, purpose-built to push the boundaries of on-device inference. Our primary contributions are as follows:

{\setlength{\leftmargini}{1em}%
\begin{itemize}
\item \textbf{X-MTP: A Lightweight Multi-Token Prediction Architecture.} We propose a shared-KV MTP design that eliminates the per-depth KV-cache replay overhead inherent in conventional MTP~\cite{deepseekai2024deepseekv3,xiao2026mimo}, substantially reducing the auxiliary parameter count and memory traffic during speculative decoding. This design is specifically optimized for the stringent latency and memory constraints of edge devices. A lightweight verification head further enables adaptive, rollback-free drafting on the hybrid linear-attention backbone.

\item \textbf{IronLLM-0.6B-Light: An Ultra-Efficient Architectural Variant.} We further explore the absolute limits of on-device efficiency by introducing IronLLM-0.6B-Light, which incorporates DyT (Dynamic Tanh)~\cite{zhu2025transformersnormalization} to pioneer the RMSNorm-free LLM architecture. Combined with a series of aggressive yet principled simplifications---including learnable upper-bounded ReLUx activations~\cite{choi2018pactparameterizedclippingactivation} and data-independent gating---this variant systematically eliminates computational bottlenecks to maximize inference throughput.

\item \textbf{A Large-Scale, Data-Efficient Processing Framework.} We develop a unified data pipeline that achieves extreme data efficiency~\cite{olmo2026olmo3}, consuming merely 6.2 trillion pre-training tokens while yielding performance competitive with models trained on substantially larger corpora. This framework integrates systematic quality enhancement, data composition optimization, and a continuous data-model co-optimization loop tailored to the heightened data sensitivity of small-parameter models.

\item \textbf{Multi-Domain On-Policy Distillation (MOPD) for Post-Training.} We employ an MOPD paradigm~\cite{ma2026mopd,agarwal2024policy,gu2024minillm} during post-training that decouples capability production from capability integration. By distilling multiple domain-specialized teachers into a unified student through on-policy distillation with verifiable rewards, we achieve robust multi-domain performance while mitigating the cross-domain interference that commonly afflicts compact models.
\end{itemize}}

Through these combined innovations, IronLLM-0.6B achieves performance on par with top-tier edge models such as Qwen3.5-0.8B~\cite{qwen3.5} and MiniCPM5-1B~\cite{minicpm4}, while its Instruct-Only design generates substantially shorter and more concise responses---a critical advantage for latency-sensitive applications. As summarized in Figure~\ref{fig:model_score_and_relative_efficiency}, the IronLLM models occupy a favorable position on the capability--efficiency frontier of compact language models: at a 32K context length, Qwen3-0.6B decodes at only 0.44$\times$ the speed of IronLLM-0.6B. Moreover, the proposed X-MTP architecture and IronLLM-0.6B-Light variant collectively establish a new paradigm for on-device efficiency, unlocking unprecedented possibilities for extreme-speed inference in future edge-centric scenarios, from autonomous robotics to next-generation intelligent cockpits.
\section{Architecture}
\label{sec:architecture}
In this section, we present the architecture of IronLLM-0.6B, together with IronLLM-0.6B-Light, a structurally streamlined variant optimized for on-chip deployment that strikes a balance between competitive performance and substantially reduced inference cost.

\subsection{IronLLM-0.6B Basic Architecture}
\begin{figure}[htbp]
    \centering
    \includegraphics[width=0.9\textwidth]{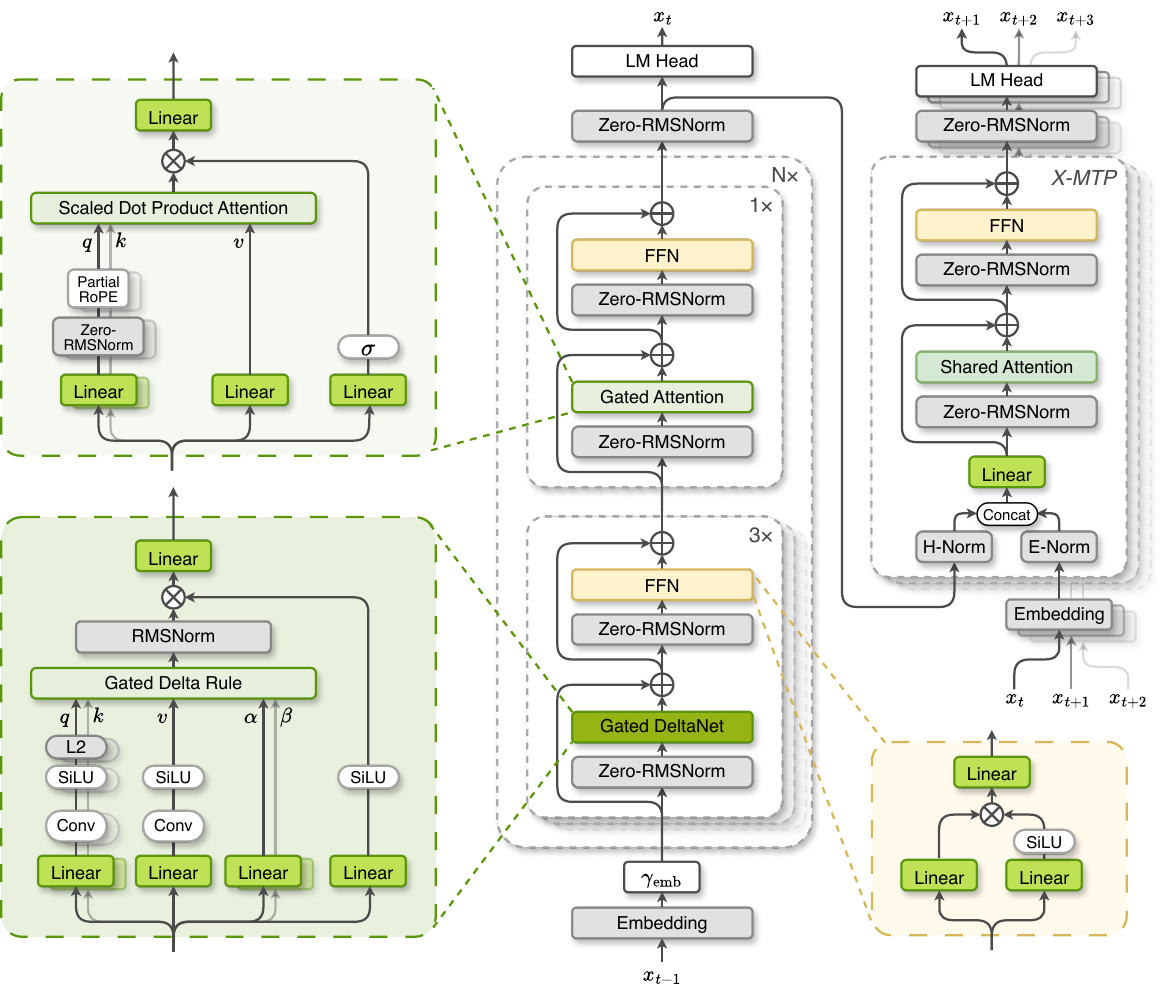}
    \caption{\textbf{Overview of the IronLLM-0.6B architecture.} Gated DeltaNet (GDN) linear-attention layers and gated global attention (GA) layers are interleaved at a 3:1 ratio, with Zero-RMSNorm adopted throughout for training stability. An X-MTP module with a shared attention block and tied embedding/LM-head weights provides multi-token prediction.}
    \label{fig:model_arch_0730}
\end{figure}

As illustrated in Figure~\ref{fig:model_arch_0730}, IronLLM-0.6B follows a design philosophy similar to Qwen3.5~\cite{qwen3.5}, adopting a decoder-only Transformer~\cite{vaswani2017attention} with a hybrid attention mechanism and a parameter budget of approximately 650M, carefully calibrated for efficient deployment on edge devices while retaining strong bilingual (Chinese--English) capabilities. The model consists of 24 Transformer blocks in total, with embedding and output projection weights tied to reduce the parameter overhead introduced by the large vocabulary.

\paragraph{Hybrid Attention.} Recent mainstream LLMs increasingly replace a fraction of full self-attention layers with linear attention variants~\cite{merrill2026olmohybridtheorypractice,kimiteam2026kimik3openfrontier,qwen3.5}, among which Kimi Delta Attention (KDA)~\cite{kimiteam2025kimilinearexpressiveefficient} and Gated DeltaNet (GDN)~\cite{yang2025gateddeltanetworksimproving} are the two most representative designs. Given edge-deployment constraints, we adopt the simpler GDN formulation as our linear attention layer. Following our early experiments and common industry practice~\cite{qwen3.5}, linear and global attention layers are interleaved at a 3:1 ratio and evenly distributed across the network depth.

Each GDN layer first applies a short causal convolution to the projected queries, keys, and values to capture local token-mixing patterns, with queries and keys further $\ell_2$-normalized to stabilize training. The layer then maintains a fixed-size recurrent state $\mathbf{S}_t \in \mathbb{R}^{d\times d}$, updated at each step via a gated delta rule:
\begin{equation}
    \mathbf{S}_t = \mathbf{S}_{t-1}\left(\alpha_t\left(\mathbf{I} - \beta_t \boldsymbol{k}_t \boldsymbol{k}_t^\top\right)\right) + \beta_t \boldsymbol{v}_t \boldsymbol{k}_t^\top, \qquad \boldsymbol{o}_t = \mathbf{S}_t \boldsymbol{q}_t,
\end{equation}
where $\alpha_t, \beta_t \in (0,1)$ are input-dependent gates controlling state decay and write strength, respectively. An additional output gate rescales $\boldsymbol{o}_t$ to compensate for the absence of softmax-style normalization.

The remaining global attention (GA) layers retain standard Grouped-Query Attention (GQA)~\cite{ainslie2023gqa} with QK-Norm~\cite{dehghani2023scalingvisiontransformers22} for precise long-range retrieval, and adopt Partial Rotary Positional Embeddings (RoPE)~\cite{su2021roformer} to further strengthen long-context extrapolation. This hybrid design substantially reduces inference cost on long sequences while preserving retrieval fidelity.

\paragraph{Training Stability.}
Following Qwen3.5~\cite{qwen3.5}, we employ output gating in both linear and global attention layers to suppress the attention and residual sinks caused by outlier activations~\cite{qiu2026unifiedviewattentionresidual}, and also adopt Zero-RMSNorm for training stability. Formally, for an input vector $\boldsymbol{x} \in \mathbb{R}^d$, Zero-RMSNorm is defined as
\begin{equation}
    \mathrm{Zero}\text{-}\mathrm{RMSNorm}(\boldsymbol{x}) = \frac{\boldsymbol{x}}{\sqrt{\frac{1}{d}\sum_{i=1}^{d}x_i^2 + \epsilon}} \odot (\mathbf{1}+\boldsymbol{\gamma}), \qquad \boldsymbol{\gamma} \leftarrow \mathbf{0},
\end{equation}
where the learnable gain $\boldsymbol{\gamma}$ is initialized to $\mathbf{0}$ rather than $\mathbf{1}$ as in conventional RMSNorm~\cite{jiang2023prermsnormprecrmsnormtransformersequivalent} to avoid the systematic scale drift.

\paragraph{Tokenizer.} Given our focus on Chinese--English bilingual scenarios, we adopt the Qwen3 tokenizer~\cite{yang2025qwen3} with a vocabulary size of 152K. Its high compression rate on multilingual text is critical for small edge-side models, where the embedding and output layers account for a substantial fraction of the total parameter count. Compared with the Qwen3.5 tokenizer, this choice reduces the parameter count by approximately 100M.

\begin{table}[ht]
  \centering
  \caption{\textbf{Core architecture configuration of IronLLM-0.6B.}}
  \label{tab:architecture}
  \small
  \begin{tabular}{lc}
    \toprule
    \textbf{Component} & \textbf{IronLLM-0.6B} \\
    \midrule
    Total parameters                & 654M    \\
    Number of layers                & 24      \\
    Hidden size                     & 1024    \\
    Intermediate size               & 3584    \\
    Attention type                  & Hybrid  \\
    Positional encoding             & Partial RoPE \\
    Number of GA layers             & 6       \\
    GA heads (Q/KV)                 & 8/2     \\
    GA head dimensions (QK/V)       & 256     \\
    Linear heads (Q/KV)             & 16/16   \\
    Linear head dimensions (QK/V)   & 128/128 \\
    Vocabulary size                 & 152K    \\
    \bottomrule
  \end{tabular}
\end{table}

Table~\ref{tab:architecture} summarizes the detailed configuration of IronLLM-0.6B. The model comprises 18 GDN layers and 6 GA layers, interleaved at a 3:1 ratio to balance linear-attention efficiency with full-attention retrieval fidelity. In total, IronLLM-0.6B contains 654M parameters, with a hidden size of 1024 and an FFN intermediate dimension of 3,584. The GA layers adopt GQA with 8 query heads and 2 key--value heads (head dimension 256), while the GDN layers employ 16 linear-attention heads (head dimension 128 for Q, K, and V). This heterogeneous head configuration, together with the stabilization designs described above, yields a compact yet expressive architecture tailored for efficient on-device inference.

\subsection{IronLLM-0.6B-Light: Edge-Oriented Model Structure}

\begin{figure}[htbp]
    \centering
    \includegraphics[width=0.8\textwidth]{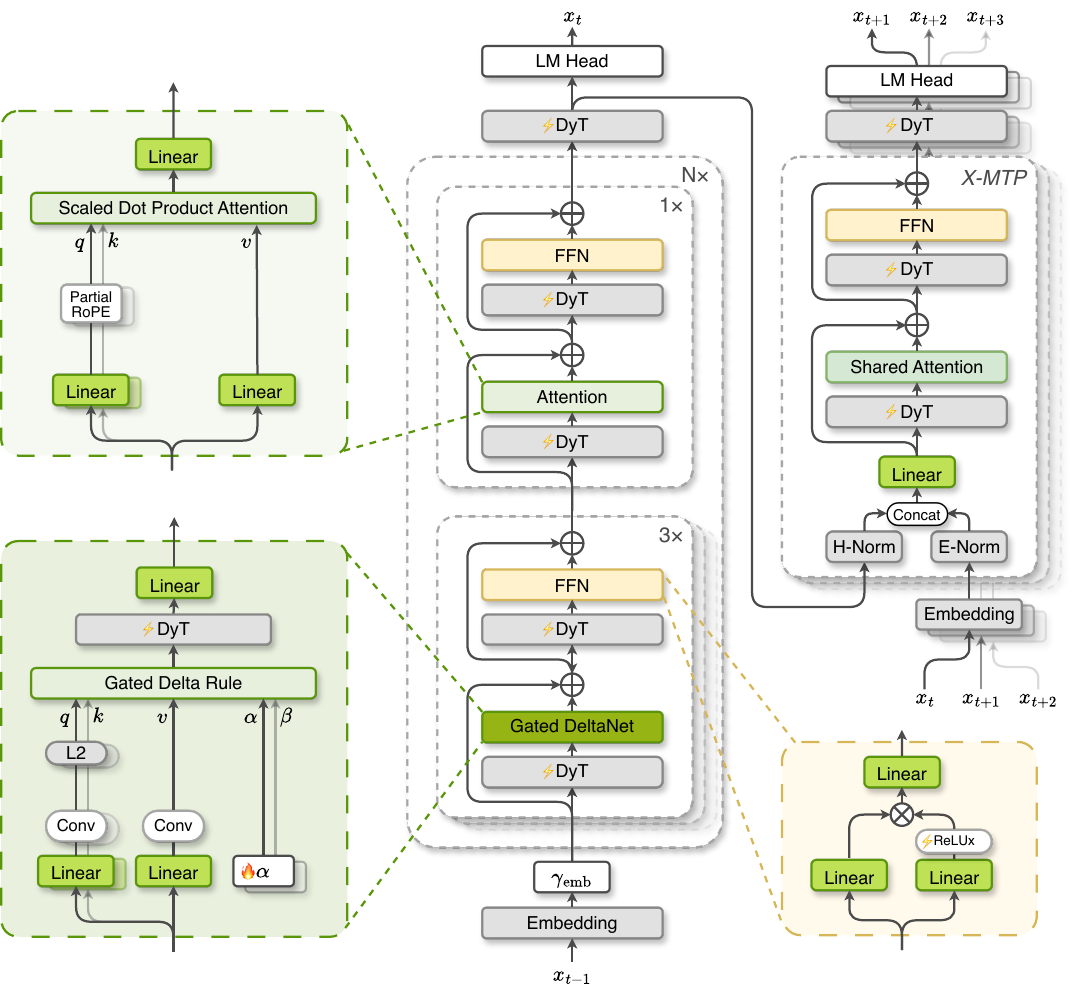}
    \caption{\textbf{Overview of the IronLLM-0.6B-Light architecture.} Building on IronLLM-0.6B, the Light variant replaces all RMSNorm layers with Dynamic Tanh (DyT), removes the attention output gate and QK-Norm, omits the SiLU activation after the causal convolution in GDN, and adopts the upper-bounded ReLUx activation, reducing inference cost and improving quantization friendliness.}
    \label{fig:model_arch_light_0804}
\end{figure}

To push the practical limits of on-device inference, we introduce IronLLM-0.6B-Light (Figure~\ref{fig:model_arch_light_0804}), a streamlined variant of IronLLM-0.6B tailored for extreme edge-side deployment, with a primary focus on maximizing inference throughput and latency efficiency while keeping performance degradation under control.

Our core design philosophy is to systematically identify and simplify modules that incur significant computational overhead or hinder quantization-friendly deployment. Specifically, we target two major bottlenecks: (1) \textit{computationally intensive RMSNorm operations}~\cite{jiang2023prermsnormprecrmsnormtransformersequivalent}, and (2) \textit{sigmoid-based activations}, which exhibit poor compatibility with post-training quantization.

\paragraph{Dynamic Tanh (DyT) for Normalization.} We replace all RMSNorm layers with Dynamic Tanh (DyT)~\cite{zhu2025transformersnormalization} operation, defined as:
\begin{equation}
    \text{DyT}(\boldsymbol{x})=
    \boldsymbol{\gamma}
    \odot
    \tanh\left(\alpha\boldsymbol{x}\right)
    +
    \boldsymbol{\beta},
\end{equation}
where $\alpha$, $\boldsymbol{\gamma}$, and $\boldsymbol{\beta}$ are learnable parameters. This substitution not only reduces computational cost but also eliminates the reliance on root-mean-square statistics, which are expensive to compute on resource-constrained devices. Additionally, we remove the QK-Norm~\cite{dehghani2023scalingvisiontransformers22} previously used in the attention module, as its functionality is largely subsumed by DyT.

\paragraph{ReLUx: Learnable Upper-Bounded ReLU.} To make the activation function friendly to on-chip deployment, we introduce ReLUx, a variant of the standard ReLU with a learnable upper bound~\cite{choi2018pactparameterizedclippingactivation}. While conventional ReLU is defined as
\begin{equation}
\text{ReLU}(x) = \max(0, x),
\end{equation}
ReLUx is formulated as
\begin{equation}
\text{ReLUx}(x) = \min(\max(0, x), \theta),
\end{equation}
where \(\theta\) is a learnable parameter, applied per layer or per head. The bounded activation improves numerical stability under quantization while preserving the nonlinear expressiveness of the original ReLU.

\paragraph{Learnable embedding scaling.}
Unlike conventional architectures, we introduce a learnable global scaling factor 
$\gamma_{\text{emb}}$ applied directly to the output of the token embedding layer:
\begin{equation}
    \boldsymbol{e}_t' = \gamma_{\text{emb}} \cdot \boldsymbol{e}_t
\end{equation}
where $\boldsymbol{e}_t$ is the raw token embedding, and $\gamma_{\text{emb}}$ is 
initialized to $\sqrt{d_{\text{model}}}$ and thereafter optimized jointly with all 
other model parameters. Empirically, we find that this simple modification helps 
our Light model achieve better convergence and more stable training.

\paragraph{Simplification of Gating and Kernel Modules.}
Our ablations on architectural variants show that several advanced design choices---while effective in larger models---yield only marginal gains at the 0.6B scale, where the capacity bottleneck imposed by the limited parameter count outweighs the benefits of more sophisticated modules. We therefore adopt a series of simplifications that reduce computational overhead with negligible performance regression.

Specifically, we first remove the output gating from the full-attention layers entirely, as its contribution proves negligible in this regime. Second, we revisit the SiLU~\cite{ramachandran2017searchingactivationfunctions} activation following the causal Conv1D kernel in Gated DeltaNet. When seeking a quantization-friendlier substitute, we find that ReLU-style activations actually degrade performance; counterintuitively, \emph{directly omitting} the activation yields better results than replacing SiLU with ReLU or ReLUx. We attribute this to ReLU-style gating suppressing a large fraction of feature channels to zero, which over-restricts the representational capacity of an already compact model. By contrast, removing the activation preserves full feature flow while reducing computation, leading to a more favorable efficiency--accuracy trade-off.

Moreover, in the Gated DeltaNet recurrence
\begin{equation}
\mathbf{S}_t
=
\alpha_t\mathbf{S}_{t-1}
+
\beta_t
\left(
\boldsymbol{v}_t
-
\alpha_t\mathbf{S}_{t-1}\boldsymbol{k}_t
\right)
\boldsymbol{k}_t^\top.
\end{equation}
the forget and write coefficients are originally \emph{data-dependent}. For each token \(\boldsymbol{x}_t\) and head \(h\), the baseline computes
\begin{align}
\beta_{t,h}&=\sigma\bigl(\mathbf{W}_b \boldsymbol{x}_t\bigr)_h,\\
g_{t,h}&=-\mathrm{e}^{A_h}\,\mathrm{softplus}\bigl((\mathbf{W}_a \boldsymbol{x}_t)_h+b^{\mathrm{dt}}_h\bigr),\qquad
\alpha_{t,h}=\mathrm{e}^{g_{t,h}},
\end{align}
where \(\mathbf{W}_a,\mathbf{W}_b\) are input projections, \(A_h\) and \(b^{\mathrm{dt}}_h\) are learnable per-head parameters, and \(\sigma\) denotes the sigmoid. This requires dynamic projection and nonlinearities at every timestep.

In our Light variant, we replace both coefficients with \textit{data-independent}, \textit{learnable head-wise} scalars, eliminating \(\mathbf{W}_a\) and \(\mathbf{W}_b\):
\begin{align}
\beta_h&=\sigma\bigl(\hat{\beta}_h\bigr),\\
g_h&=-\bigl(\mathrm{softplus}(\gamma_h)+\varepsilon\bigr),\qquad
\alpha_h=\mathrm{e}^{g_h},
\end{align}
where \(\hat{\beta}_h,\gamma_h\in\mathbb{R}\) are trainable per-head parameters (\(\hat{\beta}_h\) initialized at \(0\), hence \(\beta_h{=}0.5\)), and \(\varepsilon{=}10^{-6}\) ensures \(\alpha_h\in(0,1)\). For the decay logits \(\gamma_h\), we adopt a Lightning-Attention-style~\cite{qin2024lightningattention2, press2022alibi} slope initialization with layer-wise scaling:
\begin{equation}
r_{l,h}=s_h\cdot\Bigl(1-\frac{l}{L-1+\delta}+\delta\Bigr),\qquad
\gamma_{l,h}=\mathrm{softplus}^{-1}(r_{l,h}),
\end{equation}
where \(s_h\) is the standard power-of-two slope schedule over value heads, \(l\) is the layer index, \(L\) is the total number of layers, and \(\delta{=}10^{-5}\) is a small offset that keeps \(r_{l,h}\) strictly positive at all layers. Collectively, these changes remove token-wise dynamic gating, reduce projection cost, and yield more predictable inference behavior across varying input lengths.

\section{Training Data}
\label{sec:data_pipeline}
In this section, we present the data pipeline underlying our model training. We begin by introducing the diverse data sources used to construct the training corpus, followed by our unified data processing pipeline for large-scale data construction, curation, and optimization. Finally, we describe our benchmark contamination analysis, which helps ensure a consistent and high-quality training corpus.

\subsection{Data Sources}
We construct our bilingual training corpus through a unified data acquisition framework that continuously integrates diverse public and internally processed data sources. Rather than relying on a fixed collection of datasets, we continuously expand and refine the corpus throughout the model development lifecycle.

The training data spans a broad range of domains, including general web content, academic literature, code, mathematics, educational content, reasoning-oriented data, and other high-value sources. These data provide complementary signals for language understanding, knowledge acquisition, reasoning, mathematical problem solving, and code generation.

Beyond existing public resources, we continuously incorporate newly available data through our in-house data acquisition workflow, allowing the composition and coverage of the training corpus to evolve over time. The detailed data processing and optimization pipeline is introduced in the next section.

\subsection{Data Processing Pipeline}
\begin{figure}[htbp]
    \centering
    \includegraphics[width=1\textwidth]{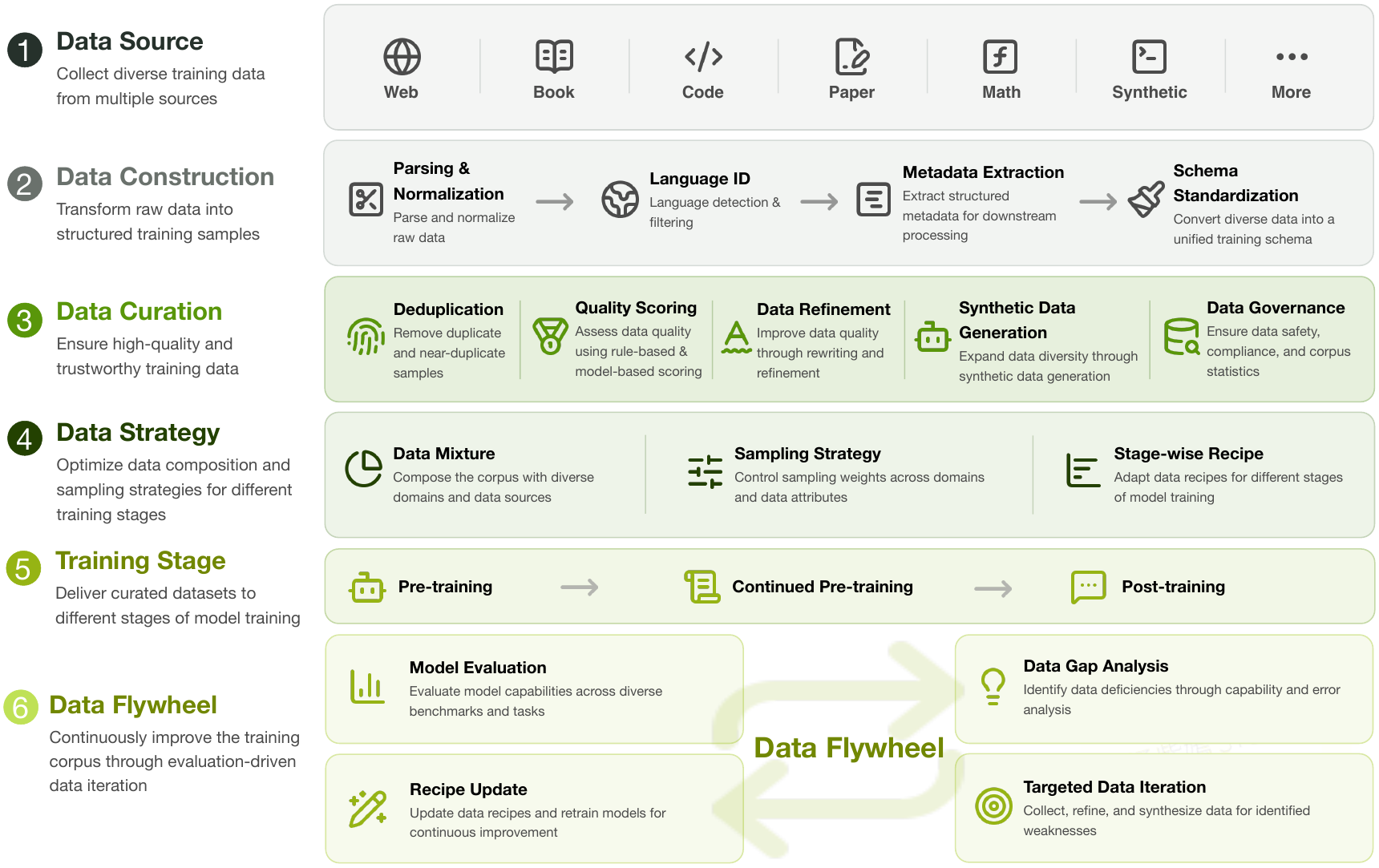}
    \caption{\textbf{Training Data Ecosystem.}}
    \label{fig:data_pipe_0730}
\end{figure}

To support scalable and high-quality model training across the entire training lifecycle, we build a unified data processing pipeline that continuously transforms raw data into a high-quality training corpus. As illustrated in Figure~\ref{fig:data_pipe_0730}, the pipeline integrates data construction, curation, strategy, and continuous data iteration into a unified framework, connecting diverse data sources with different stages of model training. Standardized data construction transforms heterogeneous raw data into structured training samples, quality-aware data curation improves the consistency and reliability of the training corpus, flexible data strategies organize stage-specific data compositions for different training objectives, and continuous data iteration enables the training corpus to evolve alongside model development through a closed-loop feedback process.

Building upon this unified data engine, we continuously improve the training corpus through three key capabilities:

\paragraph{Systematic Data Quality Enhancement.}We establish a unified data quality framework with standardized processing procedures across diverse data sources. Instead of relying on isolated heuristics, we develop in-house quality assessment models and ensemble multiple quality signals, including model-based scoring, rule-based signals, and domain-specific indicators, to achieve more comprehensive and reliable data evaluation. Based on the quality assessment results, we apply differentiated optimization strategies: low-quality samples are filtered, medium-quality samples are refined through targeted enhancement techniques, and high-quality samples are further utilized for synthetic data generation. This quality-aware data refinement process enables continuous improvement of the training corpus beyond simple data collection and filtering.

\paragraph{Data Composition Optimization.}Beyond improving individual sample quality, we optimize the overall data distribution through systematic data mixture experiments. We establish a unified domain taxonomy and perform fine-grained domain classification to guide mixture design, enabling data-driven optimization of training recipes. With extensive mixture ratio exploration, we achieve competitive model performance while consuming only approximately 6.2T pre-training tokens, demonstrating substantially improved data efficiency. More importantly, the optimized mixture strategy is not limited to a single training stage. The same recipe search methodology can be applied across different stages and capability domains, enabling efficient construction of stage-specific data mixtures.

\paragraph{Data-Model Co-optimization Loop.}We build a continuous data optimization loop that integrates model evaluation, capability analysis, and targeted data improvement. By analyzing model performance and identifying capability gaps, we further attribute these gaps to potential data deficiencies, such as insufficient coverage, imbalanced distributions, or limited high-quality supervision. Based on these insights, we develop targeted data strategies, including domain-specific data expansion, custom processing pipelines, and synthetic data generation. This iterative optimization paradigm has been applied across multiple capability domains, including Chinese language understanding, mathematical reasoning, and code generation, enabling the training corpus to continuously evolve with model development.

Together, these capabilities transform training data from a static collection into a measurable, controllable, and continuously evolving data asset, enabling long-term model improvement.

\subsection{Data Contamination Analysis}
Benchmark contamination may lead to overly optimistic evaluation results if benchmark samples appear in the training corpus. To ensure the reliability of our reported performance, we perform comprehensive contamination analysis across all major evaluation benchmarks.

Our contamination analysis combines multiple complementary detection methods, including n-gram overlap detection and embedding-based semantic retrieval, to identify potential overlaps. Based on the detected overlaps, we construct cleaned evaluation sets and compare model performance on the original and cleaned versions of the benchmarks to quantify the impact of contamination.

Across most evaluated benchmarks, performance differences between original and cleaned sets remain within 2.5 points, with several benchmarks even showing slight improvements after decontamination. This suggests that benchmark contamination has limited impact on our reported results. Detailed contamination detection procedures and experimental results are provided in Appendix~\ref{sec:data_contam}.

\section{Pre-Training}
\label{sec:pretraining}
This section details our pre-training methodology, specifically optimized for small-parameter language models targeting edge deployment. We first describe our data construction strategy, which prioritizes quality and mixture efficiency over raw scale, followed by the multi-stage training curriculum. We then present the pre-training setup, including hyperparameters and training strategy. Finally, we introduce our small-model-oriented evaluation framework and the long-context mid-training pipeline that extends the model's native context window from 4,096 to 65,536 tokens.

\subsection{Pre-Training Data Construction}
Given the inherent capacity constraints of edge-deployed, small-parameter models, our pre-training data strategy diverges substantially from mainstream large-scale LLM training practices, which often prioritize data volume within fixed compute budget. Instead, we prioritize data quality and mixture efficiency over raw scale~\cite{olmo2026olmo3}.

Building on the data pipeline described in Sec.~\ref{sec:data_pipeline}, we curate a high-quality training corpus from a pre-processed pool of nearly 20 trillion tokens. Through systematic data mixture experiments, the final pre-training consumes only \textbf{6.2 trillion tokens}, approximately $1/6$ of the 36 trillion tokens used to train Qwen3 series~\cite{yang2025qwen3}.

The curated corpus spans a diverse set of domains, including public web text, books, academic papers, mathematics, and code, complemented by high-quality synthetic data in a variety of formats (e.g., QA-style pairs). Consistent with our bilingual focus, the corpus is predominantly composed of Chinese and English text.

\subsection{Multi-Stage Pre-Training Strategy}
\begin{figure}[htbp]
    \centering
    \includegraphics[width=1\textwidth]{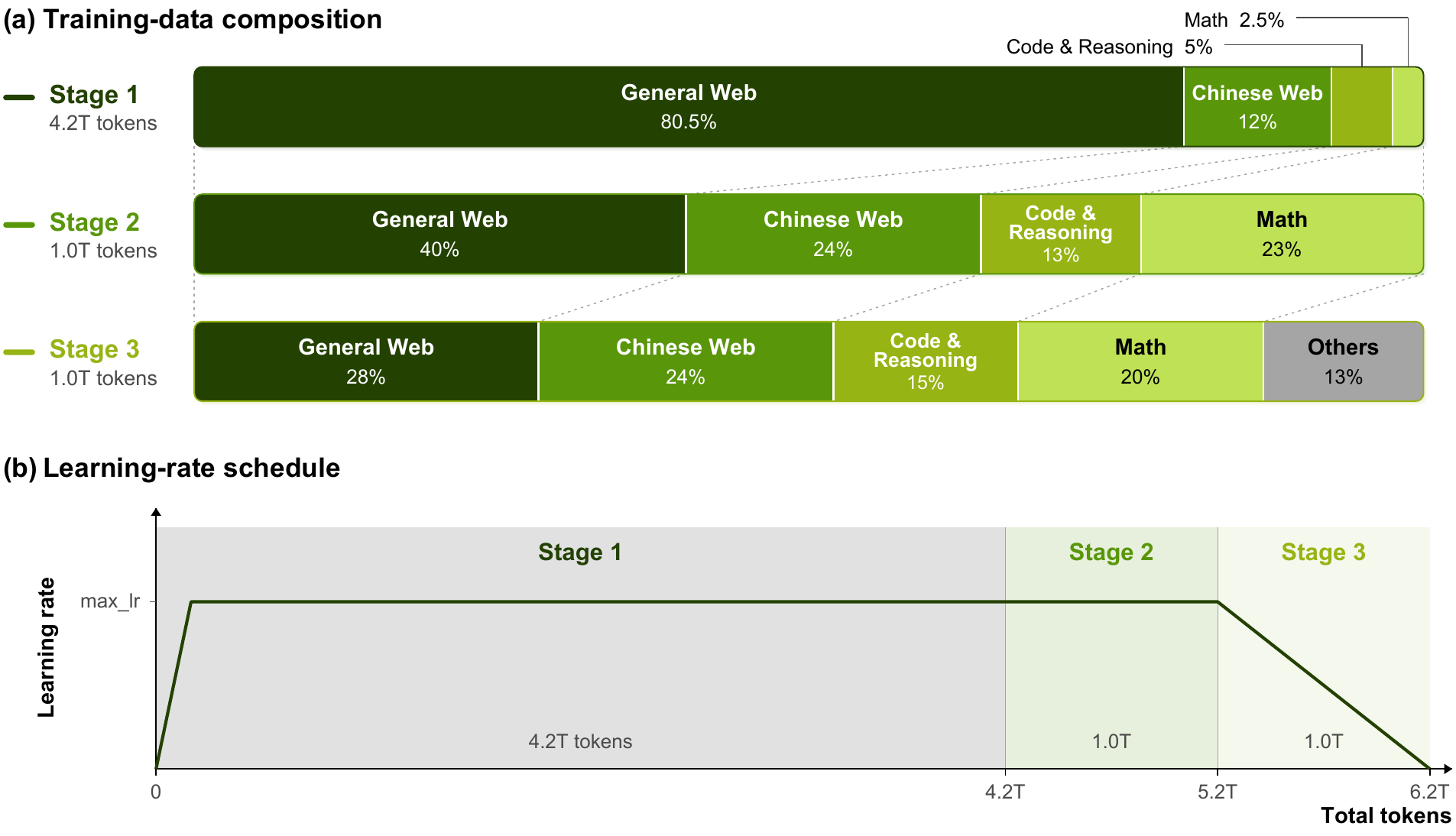}
    \caption{\textbf{Illustration of the multi-stage pre-training pipeline.} (a) Data composition across stages; (b) corresponding learning-rate schedule following the Warmup-Stable-Decay (WSD) pattern.}
    \label{fig:training_stage}
\end{figure}

We adopt a three-stage pre-training strategy with a Warmup-Stable-Decay (WSD)~\cite{hu2024minicpmunveilingpotentialsmall} learning rate schedule, as illustrated in Figure~\ref{fig:training_stage}. The data composition is progressively shifted from general web corpora toward math, code, and reasoning-intensive sources as training proceeds, while the learning rate follows a corresponding WSD trajectory across stages.

\paragraph{Stage 1.}
During the first 4.2 trillion tokens, the model is trained predominantly on general web corpora, with a small proportion of math and code data introduced early to establish foundational language understanding and seed elementary STEM reasoning capabilities. We include wiki-style encyclopedic knowledge throughout this period, but down-sample highly specialized content, such as domain-specific PDFs and arXiv papers, as our experiments indicate that small models struggle to absorb such knowledge at this point, and early exposure may even hurt general language performance.

\paragraph{Stage 2.}
Over the next 1.0 trillion tokens, we reduce the proportion of general web data and increase the share of math and code corpora to strengthen the model's logical reasoning capabilities. We also introduce a modest amount of synthetic data, including QA-style pairs, to improve instruction-following and structured-response generation.

\paragraph{Stage 3.}
In the final 1.0 trillion tokens, we further reduce general web content, keeping only the highest-quality subset after rigorous filtering, while continuing to raise the overall proportion of math, code, and high-quality synthetic data. This allows us to maximize reasoning performance within the remaining budget, using the declining learning rate to refine the model toward targeted capabilities.

Notably, even as the overall share of general web data decreases across stages, we maintain, or even increase, the proportion of Chinese web data within that category, and we carefully preserve the Chinese-data ratio within each individual data type throughout all three stages. In our experiments, high-quality Chinese data substantially improves the model's Chinese-language performance without degrading English benchmark results, underscoring the importance of balanced bilingual curation even under an aggressive, quality-driven data reduction strategy.

\subsection{Pre-Training Setup}

\paragraph{Training Hyper-parameters.} We employ the AdamW optimizer~\cite{loshchilov2019decoupledweightdecayregularization} with $\beta_1 = 0.9$, $\beta_2 = 0.95$, and a weight decay of $0.1$, which is applied exclusively to two-dimensional parameters, including embedding layers. Gradient clipping is applied with a maximum norm of $1.0$. Model parameters are initialized with a uniform range of $0.02$. The rotary positional embedding (RoPE)~\cite{su2021roformer} base theta is set to $10{,}000$ and is applied to the first $64$ channels of the head dimension.

The training process consists of three stages with distinct learning rate schedules. In Stage 1, the learning rate linearly warms up from $0$ to $4.0 \times 10^{-4}$ over the first $80$B tokens. Stage 2 maintains a constant learning rate of $4.0 \times 10^{-4}$. In Stage 3, the learning rate undergoes a linear decay from $4.0 \times 10^{-4}$ down to $0$. The global batch size is fixed at 1,024 throughout the entire pre-training process, and the training sequence length is set to 4,096 across all stages. We use FlashAttention2~\cite{dao2023flashattention2fasterattentionbetter}, Flash Linear Attention~\cite{yang2024fla} and Liger Kernel~\cite{hsu2025ligerkernel} to accelerate training.

The training objective follows the standard next-token prediction cross-entropy loss. To improve training efficiency, we initialize a single Multi-Token Prediction (MTP)~\cite{deepseekai2024deepseekv3} head for one-step-ahead prediction starting from Stage 3, with the MTP loss weight set to $0.1$.

\paragraph{Per-batch data proportioning.} We examined a training paradigm in which each global batch of 1,024 samples is constructed to exactly mirror the overall source distribution, with the goal of ensuring consistent data composition at every optimization step and thereby yielding more stable gradients. Empirically, however, this strict enforcement did not lead to improved downstream task performance. We hypothesize that the stochasticity introduced by standard global shuffling serves as an implicit regularizer: permitting per-batch ratios to fluctuate naturally facilitates broader exploration of the loss landscape and reduces the likelihood of convergence to sharp minima. Conversely, rigidly constraining per-batch composition restricts the optimization trajectory and degrades generalization. Accordingly, we enforce data proportions only at the global level over the full training run, relaxing per-batch constraints. In addition, to preserve stable gradient norms, we augment standard gradient clipping with a mechanism that skips optimizer steps when anomalous data triggers severe gradient spikes.

\paragraph{Cross-document attention.} We also evaluated cross-document attention~\cite{ding2024fewertruncationsimprovelanguage}, which prevents attention interference between distinct documents concatenated within the same sequence. Ablations across training stages showed that applying this masking during the pretraining annealing phase improved performance on some benchmarks while degrading it on others, and these discrepancies were largely neutralized after subsequent supervised fine-tuning (SFT). We therefore omit this masking during the main pretraining stage, treating unmasked document concatenation as an implicit form of data augmentation. However, we explicitly enable it during long-context training, where spurious inter-document attention in significantly longer sequences consistently degrades nearly all evaluated capabilities.

\paragraph{Model Merging.}
We apply model merging~\cite{wortsman2022model, li2025modelmergingpretraininglarge} \textit{during pretraining} to mitigate performance instability on challenging downstream tasks. By merging intermediate checkpoints, we create a more robust initialization for subsequent training stages. Additionally, this approach enables the rapid evaluation of different data mixture ratios.

As shown in Table~\ref{tab:main_results}, the merged checkpoint ($X_{\text{merge}}$) outperforms or matches both the intermediate ($X - 10\mathrm{k}$) and final ($X$) checkpoints across all benchmarks, achieving substantial gains on GSM8K (+7.51) and MATH (+5.18). This indicates that merging intermediate checkpoints effectively stabilizes pretraining and provides a stronger foundation for downstream tasks.

\begin{table}[htbp]
    \centering
    \small
    \caption{Performance comparison across benchmarks. Best results are highlighted in \textbf{bold}.}
    \begin{tabular}{lccccccc} 
        \toprule
        \textbf{Checkpoint} & \textbf{MMLU} & \textbf{CMMLU} & \textbf{CEval} & \textbf{BBH} & \textbf{GSM8K} & \textbf{MATH} & \textbf{MBPP} \\
        \midrule
        Step $X-10\mathrm{k}$ & 52.32 & 48.39 & 48.11 & 35.65 & 51.55 & 19.06 & 36.19 \\
        Step $X$              & 53.18 & 48.90 & 47.67 & 36.52 & 50.34 & 17.64 & \textbf{40.47} \\
        $X_{\text{merge}}$    & \textbf{54.17} & \textbf{49.93} & \textbf{49.49} & \textbf{37.70} & \textbf{57.85} & \textbf{22.82} & \textbf{40.47} \\
        \midrule
        Gain over Step $X$    & $+0.99$ & $+1.03$ & $+1.82$ & $+1.18$ & $+7.51$ & $+5.18$ & $+0.00$ \\
        \bottomrule
    \end{tabular}
    \label{tab:main_results}
\end{table}

\subsection{Evaluations}
\subsubsection{Evaluation Setup}
We evaluate IronLLM-0.6B-Base and IronLLM-0.6B-Light-Base against representative pretrained base models in up to one-billion parameter regime, including
Qwen3-0.6B-Base~\cite{yang2025qwen3}, Qwen3.5-0.8B-Base~\cite{qwen3.5}, and
MiniCPM5-1B-Base~\cite{minicpm4}. The evaluation aims to provide a comprehensive
comparison of small language models across general knowledge, reasoning,
Chinese language understanding, mathematical
reasoning, and code generation. All models are evaluated using a unified
evaluation pipeline~\cite{cao2026opencompassuniversalevaluationplatform} with consistent prompting, decoding, and
post-processing settings whenever applicable. \par

We organize the evaluation benchmarks into the following capability
categories:

\begin{itemize}
    \item \textbf{General Tasks:}
    MMLU (5-shot)~\cite{hendrycks2021mmlu},
    MMLU-Pro (5-shot, CoT)~\cite{wang2024mmlupro},
    MMLU-redux (5-shot)~\cite{gema2025mmluredux},
    ARC-Easy \& ARC-Challenge (0-shot)~\cite{clark2018arc},
    HellaSwag (0-shot)~\cite{zellers2019hellaswag},
    CSQA (8-shot)~\cite{talmor2019commonsenseqa},
    OBQA (0-shot)~\cite{mihaylov2018openbookqa},
    PIQA (0-shot)~\cite{bisk2020piqa},
    WinoGrande (0-shot)~\cite{sakaguchi2020winogrande},
    BBH (3-shot, CoT)~\cite{suzgun2022bbh},
    TriviaQA (5-shot)~\cite{joshi2017triviaqa}.

    \item \textbf{Mathematical Reasoning:}
    GSM8K (4-shot, CoT)~\cite{cobbe2021gsm8k},
    MATH (4-shot, CoT)~\cite{hendrycks2021math}.

    \item \textbf{Code Generation:}
    HumanEval (0-shot)~\cite{chen2021humaneval},
    MBPP (3-shot)~\cite{austin2021mbpp}.

    \item \textbf{Chinese Understanding:}
    CMMLU (5-shot)~\cite{li2023cmmlu},
    C-Eval (5-shot)~\cite{huang2023ceval},
    C$^3$ (0-shot)~\cite{sun2020c3}.
\end{itemize}

\paragraph{Small-Scale Model Evaluation.}
When evaluating the Base model, we observe that most existing benchmarks are primarily designed for large-scale models, making them overly challenging for small-parameter models (e.g., 0.6B scale). This leads to unstable performance measurements~\cite{du2025understandingemergentabilitieslanguage} and complicates fair comparisons across different architectural variants, especially for models trained from scratch. To address this, we tailor our evaluation strategy specifically for the small-model setting. During early-stage pretraining, we adopt both standard MCF (Multiple-Choice Fill) and custom-designed CF (Cloze Fill) metrics~\cite{gu2025olmesstandardlanguagemodel} to continuously track model capabilities. Furthermore, we enhance the robustness of evaluation prompts for several benchmarks through refined template designs, which helps stabilize performance trajectories throughout training. Detailed descriptions are provided in Appendix~\ref{app:base_evaluation}.

\subsubsection{Evaluation Results}
As shown in Table~\ref{tab:pretraining_evaluation}, IronLLM-0.6B-Base achieves the best performance on 8 of the 19 benchmarks, including the full MMLU suite (MMLU, MMLU-Pro, and MMLU-redux), TriviaQA, HellaSwag, WinoGrande, GSM8K, and C$^3$, while remaining on par with the best baseline on MATH.
It outperforms MiniCPM5-1B-Base on 17 of the 19 benchmarks, although the latter has over 53\% more parameters (e.g., a 21.0-point margin on GSM8K).
It also leads Qwen3.5-0.8B-Base, which is approximately 25\% larger, on 13 benchmarks and performs on par with the identically sized Qwen3-0.6B-Base, highlighting the parameter efficiency of our model.
The remaining benchmarks are led by Qwen3.5-0.8B-Base (ARC, BBH, CMMLU, and C-Eval) or Qwen3-0.6B-Base (CommonsenseQA, HumanEval, and MBPP).

We also report results for IronLLM-0.6B-Light-Base, which trades a modest amount of raw capability for improved inference and quantization efficiency. Overall, it retains most of the capabilities of IronLLM-0.6B-Base, with an average deficit of about 5.5 points across the 19 benchmarks, and remains competitive against larger baselines: it outperforms MiniCPM5-1B-Base on 15 of 19 benchmarks and leads Qwen3.5-0.8B-Base on 8 of them, offering an attractive performance--efficiency trade-off for resource-constrained deployment.

We note that MiniCPM5-1B-Base only publicly releases the checkpoint prior to its mid-training stage, and we therefore evaluate this officially released version.
Moreover, we find that continuing to increase the proportion of instruction-style data in the pre-training corpus can improve the performance of base models. However, to avoid front-loading benchmark gains that rightfully belong to post-training and, more importantly, to preserve the model's plasticity and headroom for the post-training stage, we deliberately restrain our use of such data.
As such, the results in Table~\ref{tab:pretraining_evaluation} faithfully reflect the capabilities that IronLLM-0.6B-Base acquires purely from pre-training, providing a cleaner basis for attributing the gains brought by subsequent post-training.

\subsection{Long-Context Mid-Training}
\label{sec:long_context_midtraining}
Following the main pretraining, we extend the model's native 4K context window to 64K tokens through a dedicated mid-training phase. We first describe the stage-wise extension schedule with the corresponding RoPE adaptation (Table~\ref{tab:long_context_stages}), then introduce the length-bucketed data mixing strategy and the curation of targeted synthetic data.

\paragraph{Two-stage extension with progressive RoPE scaling.}
We extend the context window from 4K to 32K and then to 64K tokens, trained on approximately 42B and 21B tokens, respectively. Since the model employs RoPE~\cite{su2021roformer}, whose relatively small base ($\theta = 10{,}000$) limits the distinguishability of distant positions, we scale the base to $10^6$ at the first stage and keep it fixed thereafter, allowing the model to adapt smoothly while preserving pretrained capabilities. Simply pushing the context length further yields no significant gains, and ablations show that allocating more tokens to the first stage brings little additional benefit, suggesting that this schedule balances training efficiency and long-context capability.

\begin{table}[t]
    \centering
    \caption{\textbf{Comparison of IronLLM-0.6B-Base with other representative base models.} \textbf{Bold} and \underline{underlined} values indicate the best and second-best non-thinking results, respectively.}
    \label{tab:pretraining_evaluation}
    \resizebox{\textwidth}{!}{
    \begin{tabular}{lcc|cc|ccc}
        \toprule
        \textbf{Benchmark}
        & \textbf{\#~Shots}
        & \textbf{Mode}
        & \makecell[c]{\textbf{IronLLM-0.6B}\\\textbf{Base}}
        & \makecell[c]{\textbf{IronLLM-0.6B}\\\textbf{-Light Base}}
        & \makecell[c]{\textbf{MiniCPM5-1B}\\\textbf{Base}\tablefootnote{The officially released MiniCPM5-1B-Base checkpoint predates mid-training stage.}}
        & \makecell[c]{\textbf{Qwen3.5-0.8B}\\\textbf{Base}}
        & \makecell[c]{\textbf{Qwen3-0.6B}\\\textbf{Base}}
        \\
        \midrule

        \rowcolor{xpgpalegreen}
        \multicolumn{8}{l}{
            \textcolor{ironblue}{\textit{\textbf{General Tasks}}}
        }
        \\
        \addlinespace[2pt]

        MMLU
        & 5-shot
        & PPL
        & \textbf{55.67}
        & 50.62
        & 45.21
        & 49.94
        & \underline{54.46}
        \\

        MMLU-Pro {\scriptsize (CoT)}
        & 5-shot
        & Gen
        & \textbf{26.45}
        & 23.24
        & 21.07
        & \underline{25.71}
        & 23.93
        \\

        MMLU-redux
        & 5-shot
        & PPL
        & \textbf{57.11}
        & 51.57
        & 46.61
        & 51.02
        & \underline{55.69}
        \\

        ARC-Challenge
        & 0-shot
        & PPL
        & 64.41
        & 61.02
        & 48.14
        & \textbf{72.54}
        & \underline{66.10}
        \\

        ARC-Easy
        & 0-shot
        & PPL
        & 79.72
        & 72.66
        & 62.26
        & \textbf{84.83}
        & \underline{83.07}
        \\

        CommonsenseQA
        & 8-shot
        & PPL
        & \underline{61.92}
        & 48.40
        & 39.64
        & 59.46
        & \textbf{63.39}
        \\

        OpenBookQA
        & 0-shot
        & PPL
        & \underline{75.00}
        & 66.02
        & 63.60
        & \textbf{75.20}
        & \textbf{75.20}
        \\

        PIQA
        & 0-shot
        & PPL
        & 71.16
        & \underline{71.22}
        & \textbf{72.52}
        & 69.15
        & 69.91
        \\

        HellaSwag
        & 0-shot
        & PPL
        & \textbf{52.65}
        & \underline{51.90}
        & 50.57
        & 49.25
        & 47.13
        \\
        
        WinoGrande
        & 0-shot
        & PPL
        & \textbf{57.30}
        & \underline{56.67}
        & 55.09
        & 55.96
        & 55.25
        \\

        BBH {\scriptsize (CoT)}
        & 3-shot
        & Gen
        & 38.84
        & 35.43
        & 38.29
        & \textbf{46.87}
        & \underline{41.52}
        \\

        TriviaQA
        & 5-shot
        & Gen
        & \textbf{31.22}
        & 27.54
        & \underline{28.55}
        & 24.53
        & 26.50
        \\

        \midrule
        \rowcolor{xpgpalegreen}
        \multicolumn{8}{l}{
            \textcolor{ironblue}{\textit{\textbf{Mathematical Reasoning}}}
        }
        \\
        \addlinespace[2pt]

        GSM8K {\scriptsize (CoT)}
        & 4-shot
        & Gen
        & \textbf{61.71}
        & 53.45
        & 40.71
        & 45.72
        & \underline{59.82}
        \\

        MATH {\scriptsize (CoT)}
        & 4-shot
        & Gen
        & \underline{31.32}
        & 22.62
        & 19.36
        & 22.64
        & \textbf{31.82}
        \\

        \midrule
        \rowcolor{xpgpalegreen}
        \multicolumn{8}{l}{
            \textcolor{ironblue}{\textit{\textbf{Code Generation}}}
        }
        \\
        \addlinespace[2pt]

        HumanEval
        & 0-shot
        & Gen
        & \underline{25.61}
        & 21.34
        & 13.41
        & 20.73
        & \textbf{28.05}
        \\

        MBPP
        & 3-shot
        & Gen
        & 30.00
        & 23.60
        & \underline{33.20}
        & 26.20
        & \textbf{37.00}
        \\

        \midrule
        \rowcolor{xpgpalegreen}
        \multicolumn{8}{l}{
            \textcolor{ironblue}{\textit{\textbf{Chinese Understanding}}}
        }
        \\
        \addlinespace[2pt]

        CMMLU
        & 5-shot
        & PPL
        & 51.62
        & 44.55
        & 35.71
        & \textbf{53.90}
        & \underline{52.00}
        \\

        C-Eval
        & 5-shot
        & PPL
        & 50.38
        & 43.80
        & 36.31
        & \textbf{55.88}
        & \underline{54.75}
        \\

        C$^3$
        & 0-shot
        & PPL
        & \textbf{57.86}
        & 50.14
        & 49.64
        & \underline{55.51}
        & 52.93
        \\

        \bottomrule
    \end{tabular}
    }
\end{table}

\paragraph{Length-bucketed data mixing.}
To strengthen long-context learning without eroding short-context performance, we partition the training corpus into length-based buckets and tune the cross-bucket sampling ratios through extensive ablations. The overall mixture otherwise closely follows that of the final pretraining stage, with only selected buckets upsampled for long-context learning. The resulting mixture consistently improves long-context benchmarks while remaining competitive on standard short-context evaluations.

\paragraph{Targeted synthetic long-context data.}
We find that capabilities such as retrieval, counting, and other specialized long-context reasoning are difficult to acquire from naturally occurring long documents alone. We therefore curate high-quality synthetic data specifically targeting these skills. Incorporating these datasets during either mid-training or SFT substantially improves the corresponding abilities without degrading general or short-context performance, making targeted synthetic data an efficient means of acquiring specialized long-context skills.

\begin{table}[htbp]
    \centering
    \caption{\textbf{Context extension schedule.} The RoPE base is scaled at Stage 1 and kept fixed at Stage 2.}
    \label{tab:long_context_stages}
    \begin{tabular}{lccc}
        \toprule
        \textbf{Stage} & \textbf{Context Window} & \textbf{RoPE Base} & \textbf{Tokens} \\
        \midrule
        Pretrain & 4,096  & 10,000                        & 6.2T \\
        Stage 1  & 32,768 & $10{,}000 \to 1{,}000{,}000$  & 42B \\
        Stage 2  & 65,536 & $1{,}000{,}000$ (fixed)       & 21B \\
        \bottomrule
    \end{tabular}
\end{table}
\section{Post-Training}
\label{sec:posttraining}

\subsection{Post-Training Pipeline}
\begin{figure}[ht]
    \centering
    \vspace{-0.5em}
    \includegraphics[width=1\linewidth]{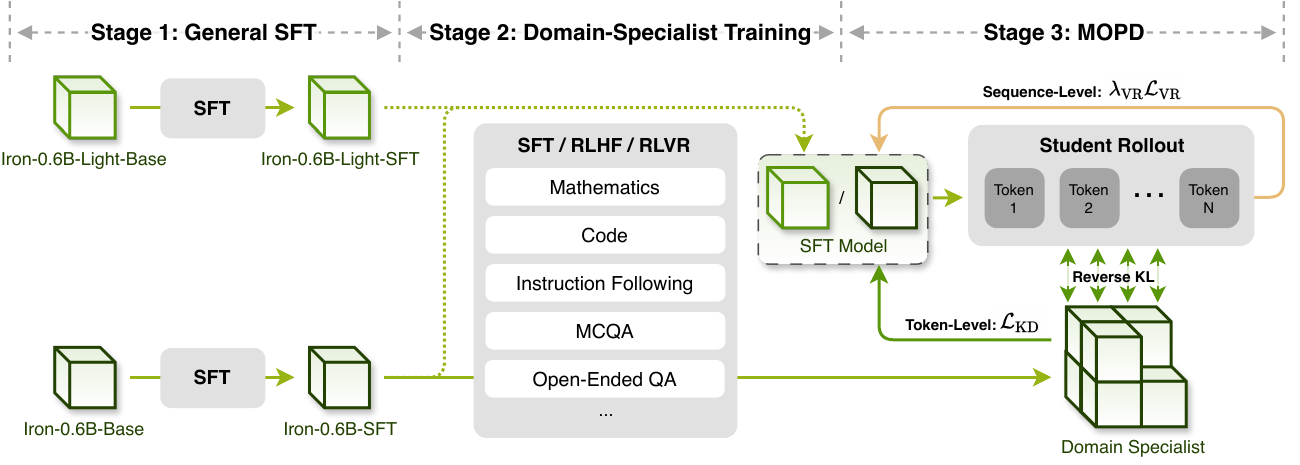}
    \vspace{-1.5em}
    \caption{\textbf{Overview of the IronLLM post-training pipeline.} IronLLM-0.6B follows three stages: General SFT, independent domain-specialist training, and Multi-Domain On-Policy Distillation (MOPD). During MOPD, frozen specialists provide token-level guidance through teacher prefill, while domain verifiers supply sequence-level verifiable rewards (VRs) on student-generated trajectories. IronLLM-0.6B-Light skips specialist training and reuses the specialists trained from IronLLM-0.6B.}
    \vspace{-1em}
    \label{fig:posttrain_pipeline}
\end{figure}

Our post-training pipeline transforms the pre-trained base model into a capable instruction-following assistant through a three-stage process, as shown in Figure~\ref{fig:posttrain_pipeline}.

\paragraph{Stage 1: General Supervised Fine-Tuning.} We first fine-tune the base model on a large-scale, multi-domain instruction dataset to establish broad conversational and instruction-following capabilities. This stage converts the base model into a general-purpose instruct model.

\paragraph{Stage 2: Domain-Specialist Training.} Starting from the General SFT model, we independently train a set of domain-specialist models, each optimized for a specific capability domain. These specialists serve as high-quality teachers in the subsequent distillation stage.

\paragraph{Stage 3: Multi-Domain On-Policy Distillation.} We distill the complementary strengths of all domain specialists back into the General SFT model via on-policy distillation, yielding a single unified model that retains broad general ability while approaching specialist-level performance in each domain.

\paragraph{Training Strategy Variants.} We apply distinct post-training strategies for different model variants to balance performance and efficiency. For IronLLM-0.6B, we execute the complete three-stage pipeline described above. For IronLLM-0.6B-Light, we streamline the process by skipping Stage 2 and directly applying MOPD after Stage 1, utilizing the domain specialists trained from IronLLM-0.6B as teachers. This design choice is motivated by three key advantages: (1) \textit{Higher Performance Ceiling}: Teachers derived from the full-capacity IronLLM-0.6B offer superior upper-bound guidance compared to specialists trained from the lightweight variant; (2) \textit{Distributional Alignment}: Since both variants share identical pre-training and General SFT stages, their output distributions remain highly similar, ensuring that cross-variant distillation remains effective without significant mismatch; and (3) \textit{Iteration Efficiency}: Bypassing the redundant training of Light-specific specialists significantly accelerates the development cycle, enabling faster experimentation and deployment.

\subsection{General Supervised Fine-Tuning}
\label{sec:SFT}

\paragraph{Data.} We curate a large-scale instruction-tuning dataset comprising several million high-quality conversation samples, encompassing both single-turn and multi-turn dialogues across diverse categories---including general QA, instruction following, mathematical reasoning, code generation, and tool-use scenarios. Notably, general QA data constitutes over one-third of the dataset, to expose the model to a wide spectrum of response formats and stylistic conventions across varied task contexts. Similarly, mathematical reasoning accounts for more than one-third of the corpus, serving as a critical foundation for cultivating robust logical reasoning capabilities; this emphasis ensures that the model can autonomously generate diverse and coherent chain-of-thought (CoT) trajectories during subsequent reinforcement learning training. This comprehensive and strategically balanced coverage guarantees that the resulting model develops strong generalist competencies prior to any domain-specific specialization.

\begin{figure}[ht]
    \centering
    \vspace{-0.5em}
    \includegraphics[width=0.4\linewidth]{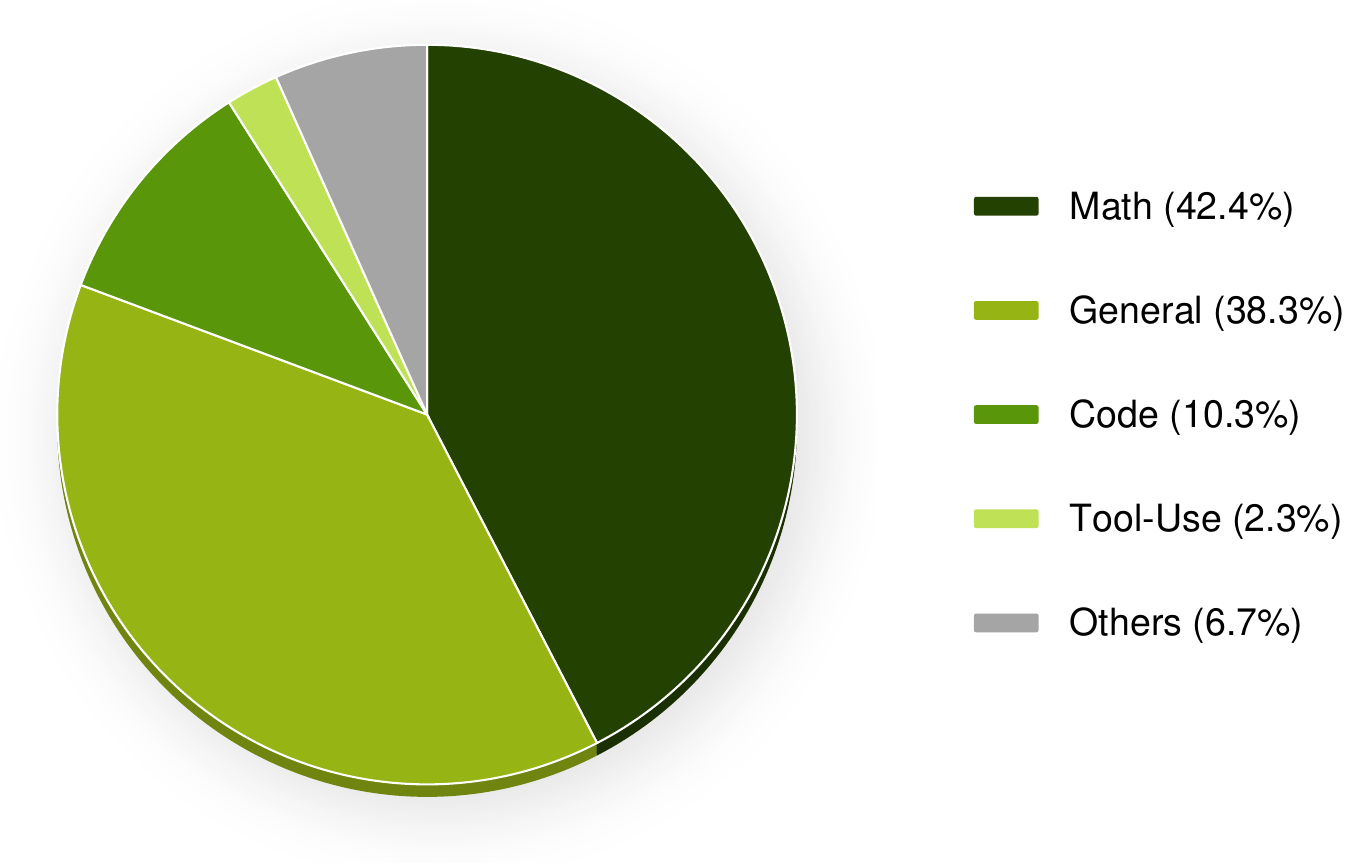}
    \vspace{-1em}
    \caption{\textbf{Composition of the general supervised fine-tuning (SFT) dataset.}}
    \label{fig:sft_data_mixture}
    \vspace{-1em}
\end{figure}

\paragraph{Training Configuration.} The model is trained for 3 epochs using the same optimizer configuration as in the pre-training stage to maintain training stability and consistency. The learning rate schedule begins with a warm-up phase over the first 100 steps, linearly increasing from zero to a peak of $8.0 \times 10^{-5}$, before smoothly decaying to zero via a cosine annealing schedule. Throughout instruction tuning, the RoPE base is kept identical to the final setting used during mid-training, preserving the positional encoding distribution to which the model has already adapted. One notable distinction from pre-training is the application of an attention mask tailored to the conversational format of SFT samples; specifically, the loss is computed exclusively on the assistant's response tokens, excluding user prompts and system instructions from the optimization objective. This alignment ensures a seamless transition from pre-training to instruction tuning, minimizing disruption to learned representations and supporting stable convergence.

\paragraph{Model Design Choice: Instruct-Only.} A notable design decision in post-training is that we train a pure instruct model without a thinking mode. While recent work~\cite{yang2025qwen3} has demonstrated that extended reasoning traces can improve performance on complex tasks, we prioritize inference efficiency given our target deployment scenario on edge/on-device platforms. Eliminating the thinking mode significantly reduces output token count and thus latency, which is critical for real-time interactive applications on resource-constrained hardware.

\subsection{Domain-Specialist Training}
Building upon the General SFT Model, we develop a suite of domain-specialist models to advance performance boundaries on specific high-value tasks. Rather than applying a uniform training protocol, each specialist adopts a tailored strategy---employing either domain-specific Supervised Fine-Tuning (SFT), Reinforcement Learning (RL), or a sequential combination of both—depending on the unique characteristics and requirements of the target domain.

\paragraph{Domain-Specific SFT.} For domains requiring precise knowledge injection or format adherence, we curate high-quality, domain-focused instruction datasets to fine-tune the General SFT model. This phase establishes robust in-domain foundational capabilities and ensures the model internalizes task-specific conventions.

\paragraph{Domain-Specific RL.} For tasks demanding advanced reasoning or nuanced preference alignment, we apply RL either as a standalone training method or as a refinement stage following SFT. This component is critical for optimizing reasoning patterns, enhancing answer correctness, and aligning outputs with complex domain-specific preferences that are difficult to capture through supervised learning alone.

Concretely, for each training prompt $x$, the model generates a set of candidate responses $\{y_1,y_2,\ldots,y_K\}\sim p_\theta(\cdot\mid x)$ via policy sampling. Each candidate is scored by a domain-specific reward signal $r(x,y_i)$, and the model parameters are updated to maximize the expected reward. We adopt Group Relative Policy Optimization (GRPO)~\cite{shao2024deepseekmath} as our RL algorithm, 
which estimates advantages within each sampled group without requiring a separate value network:
\begin{equation}
    \hat{A}_i
    =
    \frac{
        r(x,y_i)
        -
        \mathrm{mean}\left(r(x,y_j)_{j=1}^{K}\right)
    }{
        \mathrm{std}\left(r(x,y_j)_{j=1}^{K}\right)
        + \epsilon
    },
    \qquad i=1,\ldots,K,
    \label{eq:grpo-advantage}
\end{equation}
where $\epsilon>0$ is a small constant for numerical stability. The policy is then updated with a clipped surrogate objective analogous to PPO~\cite{schulman2017ppo}, applied at the token level.

The specialist family includes:
\begin{itemize}
    \item \textbf{Mathematics.} Fine-tuned on mathematical reasoning and symbolic computation datasets, followed by RL optimization against verifiable correctness rewards. During RL, the final answer is extracted from a predefined output format and evaluated by a dedicated mathematical verifier that checks the numerical or algebraic equivalence of the predicted result against the ground truth. This pipeline enables robust step-by-step problem solving and substantially reduces arithmetic and logical errors.
    \item \textbf{Code.} Trained on code completion, synthesis, and debugging corpora spanning multiple programming languages, with subsequent RL leveraging execution-based rewards. Specifically, the executable function enclosed within the code block is extracted and paired with curated test cases; the reward signal is derived from the outcomes of assertion checks on the function's inputs and expected outputs, thereby directly optimizing for functional correctness and code quality.
    \item \textbf{Instruction Following.} Further enhanced on complex, multi-constraint instruction datasets. Each training sample is annotated with a structured set of constraint conditions---such as output language, word count limits, and formatting requirements. During RL, a dedicated instruction verifier evaluates the model's response against every individual constraint, yielding a fine-grained compliance score that serves as the reward signal. This design encourages precise and reliable adherence to composite instructions.
    \item \textbf{MCQA.} Trained via RL on multiple-choice question-answering datasets that span a broad range of disciplines and subject areas, aiming to strengthen the model's general knowledge capacity. The reward is computed in a straightforward yet effective manner: the final selected option is directly extracted from the model's output and compared against the ground-truth answer, providing a binary correctness signal without the need for auxiliary verification tools.
    \item \textbf{Open-Ended QA.} Trained with a reward model that scores response quality in alignment with human preferences. The training data encompass diverse tasks such as creative writing, editing, factual question answering, and role-playing, with a controlled proportion of safety-oriented samples. For tasks demanding high factual accuracy---such as explanation, advice, and planning—we deliberately exclude content from highly specialized domains such as law and finance to mitigate the risk of generating misleading or unverifiable claims.
\end{itemize}

These specialists operate as independent teacher models and serve as the knowledge source for the subsequent distillation stage.

\subsection{Multi-Domain On-Policy Distillation}
LLM post-training typically involves integrating capabilities across multiple domains.
Existing approaches, including Mix-RL~\cite{yang2025qwen3,team2026kimi}, Cascade RL~\cite{wang2025nemotron}, Off-Policy Fine-Tuning~\cite{liu2025deepseek,zeng2025glm}, and Parameter Merging~\cite{wortsman2022model,ilharco2022editing}, provide practical means of integrating capabilities across multiple domains.
However, these approaches often suffer from cross-domain interference and capability forgetting, leading to unstable or suboptimal capability integration.

We employ Multi-Domain On-Policy Distillation (MOPD)~\cite{agarwal2024policy,gu2024minillm,ma2026mopd} to decouple capability production from capability integration.
During MOPD training, we balance data sampling across domains such that approximately the same amount of training data is routed to each domain teacher, promoting balanced capability integration across domains.
Specifically, domain-specialized teachers are trained independently to develop their respective capabilities, while a unified student integrates these capabilities through on-policy distillation and verifiable task rewards.
Let $\mathcal{D}$ denote the set of domains and $\{\pi_{\phi_d}\}_{d\in\mathcal{D}}$ the corresponding frozen teacher policies. The number of teachers is configurable and can be extended with the domain set. Given a prompt $x$, its domain metadata $d(x)$ routes the student-generated trajectory to the corresponding teacher $\pi_{\phi_{d(x)}}$. The student policy $\pi_\theta$ samples multiple responses
\begin{equation}
    y^{(i)} \sim \pi_\theta(\cdot\mid x),
    \qquad i=1,\ldots,N,
\end{equation}
and the routed teacher evaluates each response on the student-generated prefixes, providing token-level log-probabilities. In this way, domain teachers can be trained independently, while capability integration is performed on the student policy's own trajectory distribution.
For a trajectory $y=(y_1,\ldots,y_T)$, we align the student with the routed teacher by minimizing the token-level reverse KL divergence:
\begin{equation}
    \mathcal{L}_{\mathrm{RKL}}(\theta)
    =
    \mathbb{E}_{x,\,y\sim\pi_\theta}
    \left[
    \frac{1}{T}\sum_{t=1}^{T}
    D_{\mathrm{KL}}\left(
        \pi_\theta(\cdot\mid x,y_{<t})
        \,\Vert\,
        \pi_{\phi_{d(x)}}(\cdot\mid x,y_{<t})
    \right)
    \right].
    \label{eq:mopd-rkl}
\end{equation}

In practice, we estimate the reverse KL divergence over student-generated tokens using the sampled-token $k_1$ estimator.
For each generated token $y_t$, it is defined as
\begin{equation}
    \ell_{\mathrm{KD},t}
    =
    \log\pi_\theta(y_t\mid x,y_{<t})
    -
    \log\pi_{\phi_{d(x)}}(y_t\mid x,y_{<t}).
    \label{eq:mopd-k1}
\end{equation}

For training stability, the estimator is clipped and converted into a detached token-level distillation advantage:
\begin{equation}
    \hat{A}_{\mathrm{KD},t}
    =
    -\operatorname{sg}
    \left[
        \operatorname{clip}
        \bigl(\ell_{\mathrm{KD},t},-c,c\bigr)
    \right],
    \label{eq:mopd-kd-advantage}
\end{equation}
where $\operatorname{sg}[\cdot]$ denotes stop-gradient. The corresponding distillation objective is
\begin{equation}
    \mathcal{L}_{\mathrm{KD}}(\theta)
    =
    \mathcal{L}_{\mathrm{PG}}
    \left(\theta;\hat{A}_{\mathrm{KD}}\right).
    \label{eq:mopd-kd-loss}
\end{equation}

This objective provides dense token-level supervision on trajectories generated by the current student, thereby reducing the distribution mismatch associated with off-policy teacher trajectories.

Beyond teacher supervision, verifiable rewards (VRs) derived from task outcomes provide an independent and complementary learning signal. While distillation transfers domain-specific behavior from teacher policies, VRs directly indicate whether complete responses successfully solve the underlying tasks. Incorporating this signal avoids relying exclusively on policy imitation and explicitly anchors optimization to task success.
We evaluate each response using a domain-specific verifier, such as answer verification, execution-based testing, or instruction-constraint checking, and optimize the resulting VR objective alongside the distillation objective.

The overall training objective is
\begin{equation}
    \mathcal{L}_{\mathrm{total}}(\theta)
    =
    \mathcal{L}_{\mathrm{KD}}(\theta)
    +
    \lambda_{\mathrm{VR}}\mathcal{L}_{\mathrm{VR}}(\theta),
    \label{eq:mopd-total-loss}
\end{equation}
where $\lambda_{\mathrm{VR}}$ controls the relative contribution of verifiable rewards.

The distillation objective provides dense token-level guidance from the routed domain teacher, while the VR objective supplies sequence-level feedback derived from task outcomes. Together, they enable the student to integrate capabilities from multiple domain teachers while explicitly optimizing for task success.

\subsection{Evaluations}
\subsubsection{Evaluation Setup}
We evaluate IronLLM-0.6B and IronLLM-0.6B-Light against representative state-of-the-art open-source post-trained models in the sub-2B parameter regime, including Qwen3-0.6B~\cite{yang2025qwen3}, Qwen3.5-0.8B~\cite{qwen3.5}, MiniCPM5-1B~\cite{minicpm4}, and LFM2-700M~\cite{liquidai2025lfm2}. Notably, Qwen3-0.6B, Qwen3.5-0.8B, and MiniCPM5-1B are hybrid thinking models; we report their performance under both thinking and non-thinking modes to provide a comprehensive comparison, while LFM2-700M only has an instruct version. The evaluation covers seven core capability categories: general knowledge, instruction following, subjective quality, mathematics, code generation, reasoning and function calling. Furthermore, to highlight our model's advantages in inference efficiency, we present a detailed statistical analysis correlating output token length with benchmark scores.

For subjective evaluation, we adopt DeepSeek-V4-Flash-0731 as the judge model, which scores individual responses or conducts pairwise comparisons between model outputs. This LLM-as-a-Judge protocol provides complementary signals beyond objective metrics, capturing dimensions such as helpfulness, coherence, and user preference that are difficult to quantify with rule-based scoring.

The evaluation benchmarks are organized into the following capability categories:
\begin{itemize}
    \item \textbf{General Knowledge:}
    MMLU-Pro~\cite{wang2024mmlupro},
    MMLU-Redux~\cite{gema2025mmluredux},
    C-Eval~\cite{huang2023ceval}
    and CMMLU~\cite{li2023cmmlu}.

    \item \textbf{Instruction Following:}
    IFEval~\cite{zhou2023ifeval},
    IFBench~\cite{pyatkin2026ifbench}
    and Multi-IF~\cite{he2024multiif}.

    \item \textbf{Subjective Quality:}
    AlpacaEval 2.0~\cite{dubois2024alpacaevalv2}
    and ArenaHard~\cite{li2024arenahard}.
    
    \item \textbf{Mathematics:}
    MATH-500~\cite{lightman2024math500},
    GSM8K~\cite{cobbe2021gsm8k},
    AIME 2025,
    AIME 2026,
    and HMMT Feb.~2026~\cite{dekoninck2026beyond}.

    \item \textbf{Code Generation:}
    HumanEval~\cite{chen2021humaneval},
    MBPP~\cite{austin2021mbpp},
    and LiveCodeBench v6~\cite{jain2025livecodebench}.

    \item \textbf{Reasoning:}
    Big-Bench Hard~\cite{suzgun2022bbh}
    and ZebraLogic~\cite{lin2025zebralogic}.

    \item \textbf{Function Calling:}
    BFCL v3~\cite{patil2025bfcl}.

\end{itemize}

All evaluations are conducted using EvalScope v1.10.0~\cite{evalscope_2024} with 0-shot settings and default prompt templates. For high-difficulty benchmarks such as AIME, HMMT, and LiveCodeBench, we report avg@k or pass@k metrics to ensure statistical reliability, while all baseline models utilize their officially recommended sampling parameters.

\subsubsection{Evaluation Results}
As shown in Table~\ref{tab:posttrain_evaluation}, IronLLM-0.6B demonstrates particularly strong instruction following, mathematical reasoning, and subjective response quality, while also achieving competitive performance in long-context understanding and function calling. Compared with Qwen3.5-0.8B under the non-thinking setting, IronLLM-0.6B achieves consistent and often substantial improvements across nearly all benchmarks. Overall, IronLLM-0.6B also outperforms the non-thinking MiniCPM5-1B on the majority of general knowledge, instruction-following, mathematics, and function-calling benchmarks, although the latter has over 53\% more parameters, highlighting the parameter efficiency of our model. In addition, the lighter variant IronLLM-0.6B-Light retains most of these gains and remains highly competitive among models of similar scale. 
\begin{table}[ht]
    \centering
    \caption{\textbf{Comparison of IronLLM-0.6B with representative post-trained models.} \textbf{Bold} and \underline{underlined} values indicate the best and second-best non-thinking results, respectively.}
    \label{tab:posttrain_evaluation}
    \setlength{\tabcolsep}{3.5pt}
    \resizebox{\textwidth}{!}{
        \begin{tabular}{l|cc|cccc}
            \toprule
            \textbf{Benchmark} {\scriptsize\textbf{(Metric)}}
                & \makecell[c]{\textbf{IronLLM-0.6B}\tablefootnote{\label{fn:iron-generation}For the IronLLM models, we use temperature=0.7, top\_p=0.8, top\_k=-1, presence\_penalty=1.5, and repetition\_penalty=1.0. All other baseline models use their officially recommended sampling parameters.}}
                & \makecell[c]{\textbf{IronLLM-0.6B-Light}\textsuperscript{\ref{fn:iron-generation}}}
                & \makecell[c]{\textbf{Qwen3-0.6B}\tablefootnote{\label{fn:hybrid-results}Scores of Qwen3, Qwen3.5, and MiniCPM5 are reported as Non-thinking / Thinking.}}
                & \makecell[c]{\textbf{LFM2-700M}}
                & \makecell[c]{\textbf{Qwen3.5-0.8B}\textsuperscript{\ref{fn:hybrid-results}}$^{\text{,}}$\tablefootnote{We observed relatively low performance on both code and mathematics benchmarks for Qwen3.5, accompanied by severe repetition artifacts in its thinking traces across both domains. The official Qwen3.5 blog reports no code evaluation results and marks mathematics scores as ``--'', indicating that the scores are not yet available or not applicable. Our results are consistent with those reported in the MiniCPM5 blog.}}
                & \makecell[c]{\textbf{MiniCPM5-1B}\textsuperscript{\ref{fn:hybrid-results}}}
            \\
            \midrule
            \rowcolor{xpgpalegreen}
            \multicolumn{7}{l}{\textcolor{ironblue}{\textit{\textbf{Long Context}}}} \\
            \addlinespace[2pt]

            RULER\tablefootnote{RULER scores are averaged over 4K, 8K, 16K, 32K, and 64K context lengths.}
                & \textbf{87.7} & 82.1 & 40.4 / 60.0 & 56.2 & \underline{87.5} / 82.1 & 67.5 / 78.2 \\
            LongBench v2
                & 27.0 & 27.2 & \textbf{28.4} / 28.2 & 16.1 & \underline{27.8} / 25.8 & 24.3 / 26.4 \\

            \midrule
            \rowcolor{xpgpalegreen}
            \multicolumn{7}{l}{\textcolor{ironblue}{\textit{\textbf{General Knowledge}}}} \\
            \addlinespace[2pt]

            MMLU-Pro
                & \textbf{42.1} & 35.9 & 24.4 / 37.6 & 21.5 & 35.4 / 45.9 & \underline{36.8} / 47.5 \\
            MMLU-Redux
                & \textbf{60.4} & 56.2 & 46.5 / 56.0 & 47.1 & 54.2 / 63.3 & \underline{59.7} / 69.5 \\
            C-Eval
                & \underline{47.3} & 44.7 & 42.3 / 51.8 & 37.8 & 29.0 / 31.4 & \textbf{54.6} / 66.6 \\
            CMMLU
                & \underline{49.4} & 45.2 & 45.3 / 49.6 & 38.3 & 33.0 / 47.9 & \textbf{66.4} / 72.2 \\

            \midrule
            \rowcolor{xpgpalegreen}
            \multicolumn{7}{l}{\textcolor{ironblue}{\textit{\textbf{Instruction Following}}}} \\
            \addlinespace[2pt]

            IFEval
                & \textbf{75.6} & 69.5 & 56.6 / 55.1 & 62.9 & 45.3 / 57.3 & \underline{70.4} / 77.8 \\
            IFBench
                & \underline{18.0} & 16.3 & 15.7 / 15.7 & 17.0 & 15.7 / 19.7 & \textbf{32.3} / 43.2 \\
            Multi-IF\tablefootnote{Multi-IF scores are averaged across three turns using the prompt-level strict metric.}
                & \textbf{36.3} & 28.0 & \underline{30.7} / 33.1 & 26.7 & 21.8 / 32.3 & 30.2 / 41.3 \\

            \midrule
            \rowcolor{xpgpalegreen}
            \multicolumn{7}{l}{\textcolor{ironblue}{\textit{\textbf{Subjective Quality}}}} \\
            \addlinespace[2pt]

            AlpacaEval 2.0
                & \textbf{10.2} & \underline{7.1} & 3.4 / 2.6 & 7.0 & 1.6 / 2.9 & 4.0 / 3.4 \\
            ArenaHard v0.1
                & \textbf{11.9} & 6.1 & 3.5 / 5.9 & \underline{9.0} & 2.6 / 4.6 & 6.3 / 6.4 \\

            \midrule
            \rowcolor{xpgpalegreen}
            \multicolumn{7}{l}{\textcolor{ironblue}{\textit{\textbf{Mathematics}}}} \\
            \addlinespace[2pt]

            MATH-500
                & \textbf{67.4} & \underline{57.4} & 52.2 / 74.8 & 27.6 & 43.8 / 17.0 & 56.4 / 89.0 \\
            GSM8K
                & \textbf{78.6} & \underline{73.6} & 61.8 / 78.3 & 52.5 & 49.4 / 31.1 & 67.6 / 86.4 \\
            AIME 2025 {\scriptsize (Avg@16)}
                & \textbf{10.2} & 6.3 & \phantom{0}\underline{9.0} / 16.3 & 3.3 & \phantom{0}1.7 / 0.0\phantom{0} & \phantom{0}0.0 / 31.5 \\
            AIME 2026 {\scriptsize (Avg@16)}
                & \textbf{11.0} & \underline{4.4} & \phantom{0}2.1 / 12.3 & 0.2 & \phantom{0}0.0 / 0.0\phantom{0} & \phantom{0}0.2 / 36.0 \\
            HMMT Feb.~2026 {\scriptsize (Avg@16)}
                & \underline{7.6} & 5.3 & \phantom{0}3.0 / 11.0 & 0.4 & \phantom{0}1.5 / 0.0\phantom{0} & \phantom{0}\textbf{9.5} / 26.5 \\

            \midrule
            \rowcolor{xpgpalegreen}
            \multicolumn{7}{l}{\textcolor{ironblue}{\textit{\textbf{Code Generation}}}} \\
            \addlinespace[2pt]

            HumanEval
                & \underline{46.3} & 31.7 & 32.9 / 52.4 & 30.5 & 15.2 / 29.3 & \textbf{68.9} / 83.5 \\
            MBPP
                & \underline{45.0} & 42.4 & 31.8 / 39.6 & 23.4 & \phantom{0}8.4 / 14.6 & \textbf{48.8} / 72.0 \\
            LiveCodeBench v6 {\scriptsize (Pass@3)}
                & \textbf{16.6} & \underline{14.3} & 11.4 / 16.4 & 3.4 & \phantom{0}6.9 / 5.7\phantom{0} & 11.4 / 37.7 \\

            \midrule
            \rowcolor{xpgpalegreen}
            \multicolumn{7}{l}{\textcolor{ironblue}{\textit{\textbf{Reasoning}}}} \\
            \addlinespace[2pt]

            BBH {\scriptsize (3-shot)}
                & \underline{37.6} & 30.1 & 30.9 / 56.0 & 21.9 & 37.0 / 60.3 & \textbf{67.4} / 70.8 \\
            ZebraLogic
                & \underline{3.5} & 2.5 & \phantom{0}\textbf{3.8} / 29.0 & 1.2 & \phantom{0}3.3 / 23.4 & \phantom{0}1.0 / 11.2 \\

            \midrule
            \rowcolor{xpgpalegreen}
            \multicolumn{7}{l}{\textcolor{ironblue}{\textit{\textbf{Function Calling}}}} \\
            \addlinespace[2pt]

            BFCL v3
                & \textbf{49.4} & 46.5 & \underline{47.9} / 49.3 & 37.7 & 38.8 / 39.0 & 41.3 / 49.6 \\

            \bottomrule
        \end{tabular}
    }
\end{table}

\subsubsection{Inference Efficiency}
\label{sec:inference_efficiency}
As discussed in Section~\ref{sec:SFT}, we deliberately adopt a pure instruct architecture without a thinking mode, prioritizing inference efficiency for edge/on-device deployment scenarios. In this section, we provide quantitative evidence supporting the effectiveness of this design decision.

To measure how efficiently a model converts generation time into task performance, we introduce the \textit{Score Efficiency} metric, defined as:
\begin{equation}
    \text{Score Efficiency} = \frac{\text{Score} \times \text{TPS}}{\text{Average Output Token Length}},
    \label{eq:score_efficiency}
\end{equation}
where Score denotes the benchmark accuracy (or pass rate) achieved by the model, TPS denotes the decoding throughput in tokens per second measured under the same serving setup, and Average Output Token Length is the mean number of generated tokens across all evaluation samples on the corresponding benchmark. Equivalently, since Average Output Token Length divided by TPS corresponds to the average generation time per sample, Score Efficiency can be interpreted as the task score delivered per unit of wall-clock generation time. A higher Score Efficiency indicates that the model achieves comparable or superior task performance with fewer output tokens and higher decoding speed, which directly translates to lower inference latency and reduced computational cost during serving.

The GDN layers greatly boost the decoding speed of IronLLM over long generations. Table~\ref{tab:speed_longgen} measures single-user decoding speed while each model generates 32K tokens (vLLM, RTX~4090, batch size 1, no speculative decoding). IronLLM-0.6B decodes at 460 tokens/s at 1K tokens and still at 377 tokens/s at 32K, a 1.22$\times$ increase in per-token latency over the whole generation, and is 8--9\% faster than its architectural counterpart Qwen3.5-0.8B at every length. The full-attention Qwen3-0.6B starts at a similar speed but slows by 2.87$\times$ and falls to 166 tokens/s at 32K, making IronLLM 2.27$\times$ faster there. The shallower LFM2-700M (16 layers) remains about 10\% faster than IronLLM throughout; with X-MTP (Section~\ref{sec:xmtp_speed}), IronLLM exceeds it on mathematics and code.

\begin{table}[ht]
    \centering
    \caption{\textbf{Single-user decoding speed over long generations.} Decode speed
    (tokens/s) at position $L$ of a 32K-token generation at batch size 1, i.e.\ the inverse of the
    per-token latency (TPOT) averaged over the 512 tokens preceding $L$, and the growth of TPOT
    from 1K to 32K. RTX~4090, vLLM~0.17.1, BF16,
    CUDA graphs, greedy decoding; median of 3 prompts. No speculative decoding.}
    \label{tab:speed_longgen}
    \small
    \begin{tabular}{ll|*{4}{>{\centering\arraybackslash}p{1cm}}|c}
        \toprule
        \multirow{2}{*}{\textbf{Model}} & \multirow{2}{*}{\textbf{Attention}}
        & \multicolumn{4}{c|}{\makecell{\textbf{Decode speed at position $L$ (tokens/s)}}}
        & \multirow{2}{*}{\makecell{\textbf{TPOT growth}\\\textbf{1K$\to$32K}}} \\
        & & 1K & 4K & 16K & 32K & \\
        \midrule
        IronLLM-0.6B & GDN hybrid & 460 & 445 & 410 & 377 & 1.22$\times$ \\
        IronLLM-0.6B-Light & GDN hybrid & 483 & 465 & 428 & 392 & 1.23$\times$ \\
        Qwen3.5-0.8B & GDN hybrid & 422 & 409 & 380 & 352 & 1.20$\times$ \\
        LFM2-700M & Conv hybrid & 508 & 498 & 457 & 416 & 1.22$\times$ \\
        Qwen3-0.6B & Full attention & 476 & 402 & 251 & 166 & 2.87$\times$ \\
        MiniCPM5-1B & Full attention & 368 & 360 & 322 & 279 & 1.32$\times$ \\
        \bottomrule
    \end{tabular}
\end{table}

Table~\ref{tab:inference_efficiency} reports the average output token length and Score Efficiency for each model across all evaluation benchmarks.
Note that Qwen3-0.6B, Qwen3.5-0.8B, LFM2-700M, and MiniCPM5-1B only show results in non-thinking modes, as explicit thinking-mode generation is inherently at odds with the goal of inference efficiency. Under this comparison, IronLLM-0.6B demonstrates a substantial advantage, achieving the highest Score Efficiency across nearly all benchmarks by generating significantly shorter outputs while maintaining competitive accuracy, which directly validates our Instruct-Only design choice that prioritizes concise, direct responses over verbose reasoning traces. MiniCPM5-1B shows moderate efficiency but still trails IronLLM by a notable margin, while both Qwen models exhibit the lowest Score Efficiency due to residual over-generation behavior inherited from their thinking-oriented training, producing disproportionately long outputs even without explicit chain-of-thought prompting.

\subsubsection{MOPD Capability Integration}
We compare the General SFT checkpoint, independently optimized Domain Experts, a TIES parameter-merging baseline, and the unified model trained with MOPD. The Domain Experts serve as task-specific reference points, whereas TIES and MOPD consolidate capabilities from all specialists into a single model.

As shown in Table~\ref{tab:MOPD-progression}, MOPD improves over General SFT and outperforms TIES Merging on all 13 benchmarks, demonstrating consistent integration without the negative transfer observed under parameter merging. It also matches or exceeds the corresponding Domain Expert on MMLU-Pro, CMMLU, and ArenaHard, and approaches expert performance on several instruction-following and knowledge benchmarks. 
Although gaps remain on some mathematics and code tasks, the overall results show that MOPD effectively integrates complementary specialist capabilities into a unified model.

\begin{table}[t]
    \centering
    \caption{\textbf{Inference-efficiency comparison between IronLLM-0.6B and representative models.}
    A.T. and S.E. denote average output token length and Score Efficiency, respectively.}
    \label{tab:inference_efficiency}
    \small
    \setlength{\tabcolsep}{4pt}
    \resizebox{\textwidth}{!}{%
        \begin{tabular}{
            l|cc|
            >{\centering\arraybackslash}m{3.8em}
            >{\centering\arraybackslash}m{3.8em}|
            cc|cc|cc|cc
        }
            \toprule
            \multirow{2}{*}{\textbf{Benchmark {\scriptsize (Metric)}}}
            & \multicolumn{2}{c|}{\textbf{IronLLM-0.6B}}
            & \multicolumn{2}{c|}{\textbf{IronLLM-0.6B-Light}}
            & \multicolumn{2}{c|}{\textbf{Qwen3-0.6B}\tablefootnote{\label{fn:efficiency-nonthinking}The Qwen3, Qwen3.5, LFM2, and MiniCPM5 results areobtained in non-thinking mode.}}
            & \multicolumn{2}{c|}{\textbf{LFM2-700M}\textsuperscript{\ref{fn:efficiency-nonthinking}}}
            & \multicolumn{2}{c|}{\textbf{Qwen3.5-0.8B}\textsuperscript{\ref{fn:efficiency-nonthinking}}}
            & \multicolumn{2}{c}{\textbf{MiniCPM5-1B}\textsuperscript{\ref{fn:efficiency-nonthinking}}} \\
            & A.T. & S.E.
            & A.T. & S.E.
            & A.T. & S.E.
            & A.T. & S.E.
            & A.T. & S.E.
            & A.T. & S.E. \\
            \midrule

            \rowcolor{xpgpalegreen}
            \multicolumn{13}{l}{\textcolor{ironblue}{\textit{\textbf{Long Context}}}} \\
            \addlinespace[2pt]
            RULER
                & 47.0 & \textbf{703.71} & 49.3 & 652.56 & 125.1 & 53.57 & 50.4 & 463.96 & 44.3 & \underline{694.94} & 117.0 & 161.06 \\
            LongBench v2
                & 173.9 & \textbf{58.62} & 201.8 & \underline{52.91} & 868.1 & 5.44 & 1371.6 & 1.69 & 652.7 & 15.01 & 1064.1 & 4.27 \\

            \midrule
            \rowcolor{xpgpalegreen}
            \multicolumn{13}{l}{\textcolor{ironblue}{\textit{\textbf{General Knowledge}}}} \\
            \addlinespace[2pt]
            MMLU-Pro
                & 345.1 & \underline{46.01} & 401.5 & 35.01 & 36.9 & \textbf{109.68} & 675.7 & 13.26 & 4652.2 & 2.68 & 1731.3 & 5.93 \\
            MMLU-Redux
                & 177.0 & \textbf{128.54} & 212.9 & \underline{103.46} & 103.6 & 74.46 & 336.3 & 58.25 & 2225.3 & 8.57 & 385.2 & 43.24 \\
            C-Eval
                & 13.8 & \textbf{1293.00} & 137.3 & 127.68 & 7.3 & \underline{961.21} & 562.3 & 27.98 & 1311.9 & 7.77 & 1027.2 & 14.83 \\
            CMMLU
                & 86.4 & \textbf{215.60} & 126.3 & \underline{140.41} & 182.9 & 41.14 & 392.1 & 40.62 & 1476.4 & 7.87 & 897.6 & 20.63 \\

            \midrule
            \rowcolor{xpgpalegreen}
            \multicolumn{13}{l}{\textcolor{ironblue}{\textit{\textbf{Instruction Following}}}} \\
            \addlinespace[2pt]
            IFEval
                & 265.5 & \textbf{107.35} & 269.6 & \underline{101.05} & 1130.6 & 8.30 & 1560.7 & 16.75 & 334.2 & 47.70 & 806.1 & 24.38 \\
            IFBench
                & 665.0 & \textbf{10.22} & 783.7 & 8.17 & 1080.3 & 2.40 & 1329.8 & 5.32 & 672.7 & \underline{8.19} & 2992.2 & 3.01 \\
            Multi-IF
                & 226.7 & \textbf{60.33} & 358.6 & 30.58 & 522.6 & 9.74 & 705.3 & 15.75 & 368.6 & \underline{38.23} & 2743.2 & 4.95 \\

            \midrule
            \rowcolor{xpgpalegreen}
            \multicolumn{13}{l}{\textcolor{ironblue}{\textit{\textbf{Subjective Quality}}}} \\
            \addlinespace[2pt]
            AlpacaEval 2.0
                & 510.2 & \textbf{7.53} & 555.6 & 5.00 & 586.5 & 0.95 & 502.6 & \underline{5.81} & 772.8 & 0.73 & 1028.4 & 1.08 \\
            ArenaHard
                & 1058.7 & \textbf{4.25} & 1231.9 & 1.94 & 1064.4 & 0.54 & 1016.0 & \underline{3.68} & 1981.1 & 0.46 & 2650.3 & 0.66 \\

            \midrule
            \rowcolor{xpgpalegreen}
            \multicolumn{13}{l}{\textcolor{ironblue}{\textit{\textbf{Mathematics}}}} \\
            \addlinespace[2pt]
            MATH-500
                & 1029.4 & \textbf{24.68} & 1865.9 & 12.06 & 714.2 & 12.13 & 873.2 & \underline{13.15} & 8686.4 & 1.77 & 5727.6 & 2.75 \\
            GSM8K
                & 360.8 & \textbf{82.15} & 400.7 & \underline{72.02} & 319.5 & 32.10 & 417.4 & 52.36 & 3884.8 & 4.47 & 681.7 & 27.65 \\
            AIME 2025 {\scriptsize (Avg@16)}
                & 2949.3 & \textbf{1.31} & 3864.0 & 0.63 & 1611.0 & \underline{0.92} & 2019.5 & 0.69 & 14074.7 & 0.042 & 8198.9 & 0.00 \\
            AIME 2026 {\scriptsize (Avg@16)}
                & 3398.0 & \textbf{1.22} & 5304.8 & \underline{0.32} & 1596.4 & 0.22 & 1996.3 & 0.044 & 14155.3 & 0.00 & 12720.9 & 0.005 \\
            HMMT Feb.\ 2026 {\scriptsize (Avg@16)}
                & 2239.6 & \textbf{1.28} & 4115.9 & 0.50 & 1670.3 & 0.30 & 1375.9 & 0.11 & 12558.7 & 0.043 & 4673.2 & \underline{0.57} \\

            \midrule
            \rowcolor{xpgpalegreen}
            \multicolumn{13}{l}{\textcolor{ironblue}{\textit{\textbf{Code Generation}}}} \\
            \addlinespace[2pt]
            HumanEval
                & 101.6 & \textbf{171.95} & 125.2 & 76.36 & 155.4 & 35.18 &86.8 & \underline{146.13} & 841.6 & 6.37 & 827.9 & 23.22 \\
            MBPP
                & 248.9 & \underline{68.16} & 133.1 & \textbf{117.81} & 433.9 & 12.17 & 658.0 & 14.79 & 848.3 & 3.49 & 4140.7 & 3.29 \\
            LiveCodeBench v6 {\scriptsize (Pass@3)}
                & 287.4 & \textbf{21.74} & 357.4 & \underline{13.79} & 1348.5 & 1.41 & 503.1 & 2.84 & 13512.2 & 0.18 & 3870.6 & 0.82 \\

            \midrule
            \rowcolor{xpgpalegreen}
            \multicolumn{13}{l}{\textcolor{ironblue}{\textit{\textbf{Reasoning}}}} \\
            \addlinespace[2pt]
            BBH
                & 58.3 & \textbf{243.14} & 88.4 & \underline{133.43} & 85.8& 59.76 & 198.7 & 45.83 & 2580.0 & 5.05 & 680.5 & 27.62 \\
            ZebraLogic
                & 512.0 & \textbf{2.58} & 1339.5 & 0.73 & 478.3 & \underline{1.32} & 1826.1 & 0.27 & 15448.8 & 0.075 & 19148.7 & 0.015 \\

            \midrule
            \rowcolor{xpgpalegreen}
            \multicolumn{13}{l}{\textcolor{ironblue}{\textit{\textbf{Function Calling}}}} \\
            \addlinespace[2pt]
            BFCL v3
                & 381.4 & 48.86 & 541.8 & 33.67 & 81.7 & 97.28 & 61.5 & \textbf{255.28} & 60.7 & \underline{224.83} & 51.6 & 223.15 \\

            \bottomrule
        \end{tabular}
    }
    \vspace{-1em}
\end{table}

\begin{table}[!ht]
\centering
\caption{\textbf{Capability integration results for IronLLM-0.6B.} Domain Expert denotes the independently trained specialist for each domain, while TIES Merging and MOPD produce unified multi-domain models. \textbf{Bold} and \underline{underlined} values indicate the best and second-best results, respectively.}
\label{tab:MOPD-progression}
\small
\resizebox{\textwidth}{!}{
\begin{tabular}{ll|cc|cc}
\toprule
\textbf{Category}
& \textbf{Benchmark}
& \textbf{General SFT}
& \textbf{Domain Expert}
& \textbf{TIES Merging}
& \textbf{MOPD} \\
\midrule

\multirow{2}{*}{Mathematics}
& GSM8K
& 61.0 & \textbf{83.9} & 68.9 & \underline{78.6} \\
& AIME 2026 {\scriptsize (Avg@16)}
& 0.6 & \textbf{14.0} & 0.6 & \underline{11.0} \\
\midrule

\multirow{3}{*}{Code}
& HumanEval
& 37.8 & \textbf{70.1} & 13.4 & \underline{46.3} \\
& MBPP
& 25.0 & \textbf{57.8} & 43.8 & \underline{45.0} \\
& LiveCodeBench v6 {\scriptsize (Pass@3)}
& 10.9 & \textbf{24.0} & 14.9 & \underline{16.6} \\
\midrule

\multirow{2}{*}{Instruction Following}
& IFEval
& 45.8 & \textbf{77.5} & 38.5 & \underline{75.6} \\
& IFBench
& 11.6 & \textbf{19.1} & 16.0 & \underline{18.0} \\
\midrule

\multirow{4}{*}{MCQA}
& MMLU-Pro
& 31.2 & \underline{41.5} & 28.7 & \textbf{42.1} \\
& MMLU-Redux
& 54.7 & \textbf{60.6} & 50.3 & \underline{60.4} \\
& C-Eval
& 43.3 & \textbf{49.6} & 36.2 & \underline{47.3} \\
& CMMLU
& \underline{44.5} & 37.5 & 36.1 & \textbf{49.4} \\
\midrule

\multirow{2}{*}{Open-Ended QA}
& AlpacaEval 2.0
& 1.5 & \textbf{13.5} & 1.4 & \underline{10.2} \\
& ArenaHard
& 1.2 & \underline{11.0} & 2.3 & \textbf{11.9} \\

\bottomrule
\end{tabular}
}
\end{table}

\section{Lightweight Multi-Token Prediction}

\subsection{X-MTP Design}
\label{sec:x_mtp}
Inspired by recent advances in Multi-Token Prediction (MTP), including DeepSeek-V3 and MiMo-V2-Flash~\citep{deepseekai2024deepseekv3,xiao2026mimo}, we investigate lightweight MTP designs for accelerating the autoregressive decoding of IronLLM-0.6B in on-device scenarios. We consider both standard sequential MTP and a shared-KV variant, as described below.

\subsubsection{Standard MTP Architecture}
\label{sec:standard_mtp}
\begin{figure}[ht]
    \centering
    \includegraphics[width=0.9\textwidth]{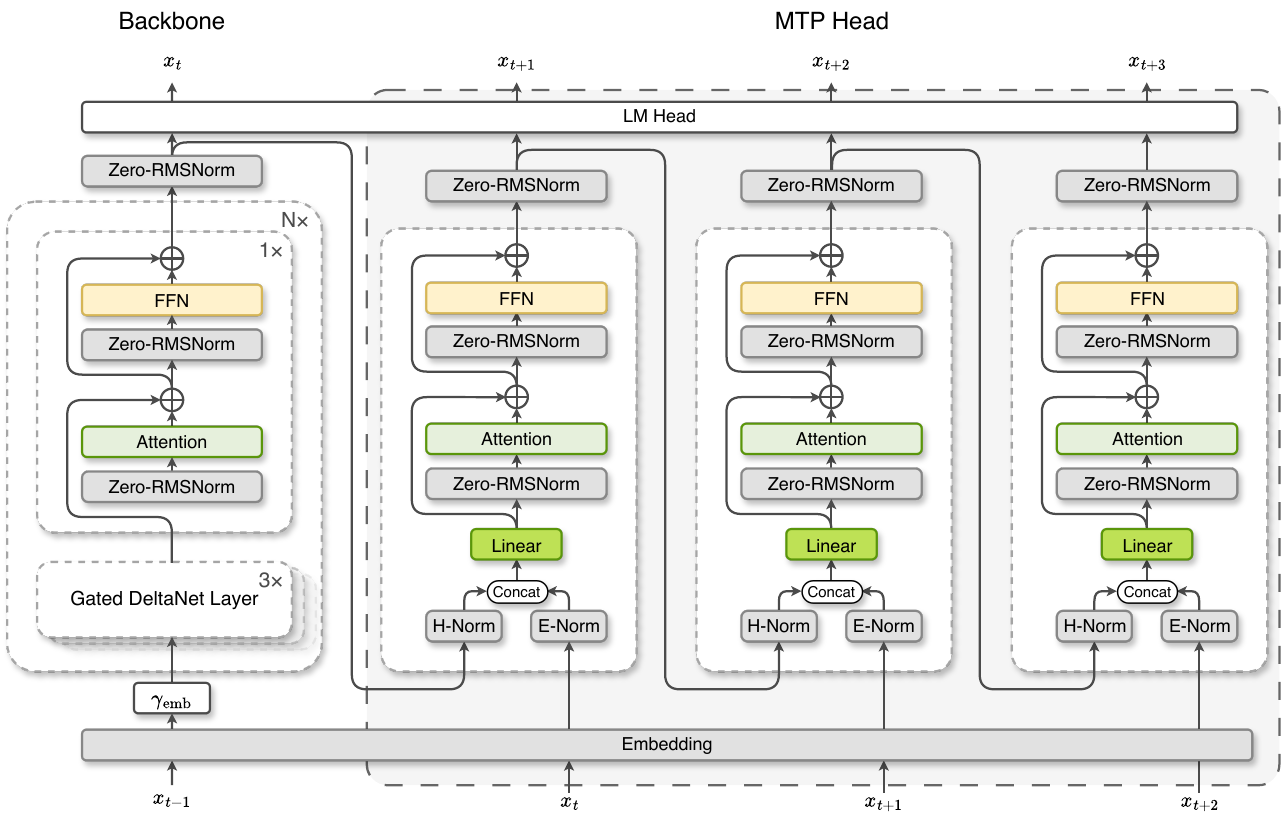}
    \caption{\textbf{The conventional MTP architecture.} Each prediction depth uses an independent MTP block.}
    \label{fig:ds_mtp_arch_0730}
\end{figure}
Let $x_{1:T}=(x_1,\ldots,x_T)\in\mathcal{V}^T$ denote an input sequence and $\boldsymbol{h}_t^{(0)} = \mathcal{F}(x_{\leq t})$ the final hidden state produced by the backbone at position $t$. As shown in Figure~\ref{fig:ds_mtp_arch_0730}, a standard MTP module predicts future tokens through a chain of auxiliary Transformer blocks. At prediction depth $k \in \{1,\ldots,K\}$, the embedding of the teacher-forced token $x_{t+k}$ is fused with the hidden state from the preceding depth:
\begin{equation}
    \boldsymbol{z}_t^{(k)} =
    \mathbf{W}_{\mathrm{fuse}}^{(k)}
    \left[
        \mathrm{Norm}_{e}^{(k)}\!\left(\mathbf{E}(x_{t+k})\right);
        \mathrm{Norm}_{h}^{(k)}\!\left(\boldsymbol{h}_t^{(k-1)}\right)
    \right],
    \qquad
    \boldsymbol{h}_t^{(k)} = \mathcal{D}^{(k)}\!\left(\boldsymbol{z}_{\leq t}^{(k)}\right),
\end{equation}
where $\mathbf{E}$ is the token embedding, $\mathcal{D}^{(k)}$ is the $k$-th MTP Transformer block, and $[\cdot;\cdot]$ denotes concatenation. The output distribution
\begin{equation}
    \boldsymbol{p}_t^{(k)}
    =
    \mathrm{softmax}\!\left(
        \mathbf{W}_{\mathrm{out}}\,\mathrm{Norm}_{o}^{(k)}
        \left(\boldsymbol{h}_t^{(k)}\right)
    \right)
\end{equation}
is supervised by $x_{t+k+1}$. Thus, the first MTP depth consumes $\mathbf{E}(x_{t+1})$ and predicts $x_{t+2}$, the second consumes $\mathbf{E}(x_{t+2})$ and predicts $x_{t+3}$, and so forth. IronLLM shares both $\mathbf{E}$ and $\mathbf{W}_{\mathrm{out}}$ with the backbone, avoiding two vocabulary-sized parameter matrices. The auxiliary objective is
\begin{equation}
    \mathcal{L}_{\mathrm{MTP}}
    =
    \frac{
    \sum_{k=1}^{K}\sum_{t}
    m_t^{(k)}
    \mathrm{CE}\!\left(
    \boldsymbol{p}_t^{(k)},x_{t+k+1}
    \right)
    }{
    \sum_{k=1}^{K}\sum_{t}m_t^{(k)}
    },
    \qquad
    \mathcal{L}
    =
    \mathcal{L}_{\mathrm{AR}}
    +
    \lambda_{\mathrm{MTP}}\mathcal{L}_{\mathrm{MTP}},
    \label{eq:mtp_loss}
\end{equation}
where $m_t^{(k)}\in\{0,1\}$ denotes whether the target
$x_{t+k+1}$ is valid. Positions affected by padding,
packed-sequence boundaries, or sequence tails are masked
out by setting $m_t^{(k)}=0$.

Besides enabling speculative decoding, this objective encourages the backbone representation to encode information about a longer prediction horizon. However, standard sequential MTP may exhibit a train-inference discrepancy. During training, the module at depth $k$ is conditioned on the ground-truth future token $x_{t+k}$ through teacher forcing. During decoding, by contrast, deeper MTP steps consume draft tokens generated by the preceding steps. An inaccurate early draft can therefore shift the inputs of subsequent steps away from their training distribution and propagate errors through the prediction chain, reducing the acceptance probability at deeper depths. This discrepancy does not negate the advantages of standard MTP, which remains an effective drafting approach, but motivates us to explore a complementary architecture.

\subsubsection{Cross-Step Parameter and KV Sharing}
\label{sec:shared_kv_mtp}

Inspired by GLM-5.2~\citep{zeng2026glm}, we further investigate shared-KV MTP, which anchors all prediction depths to the K/V states constructed at the first MTP step while retaining depth-dependent queries. Our experiments suggest that this design provides a favorable trade-off between robust multi-step drafting and inference efficiency.

\begin{figure}[ht]
    \centering
    \includegraphics[width=0.9\textwidth]{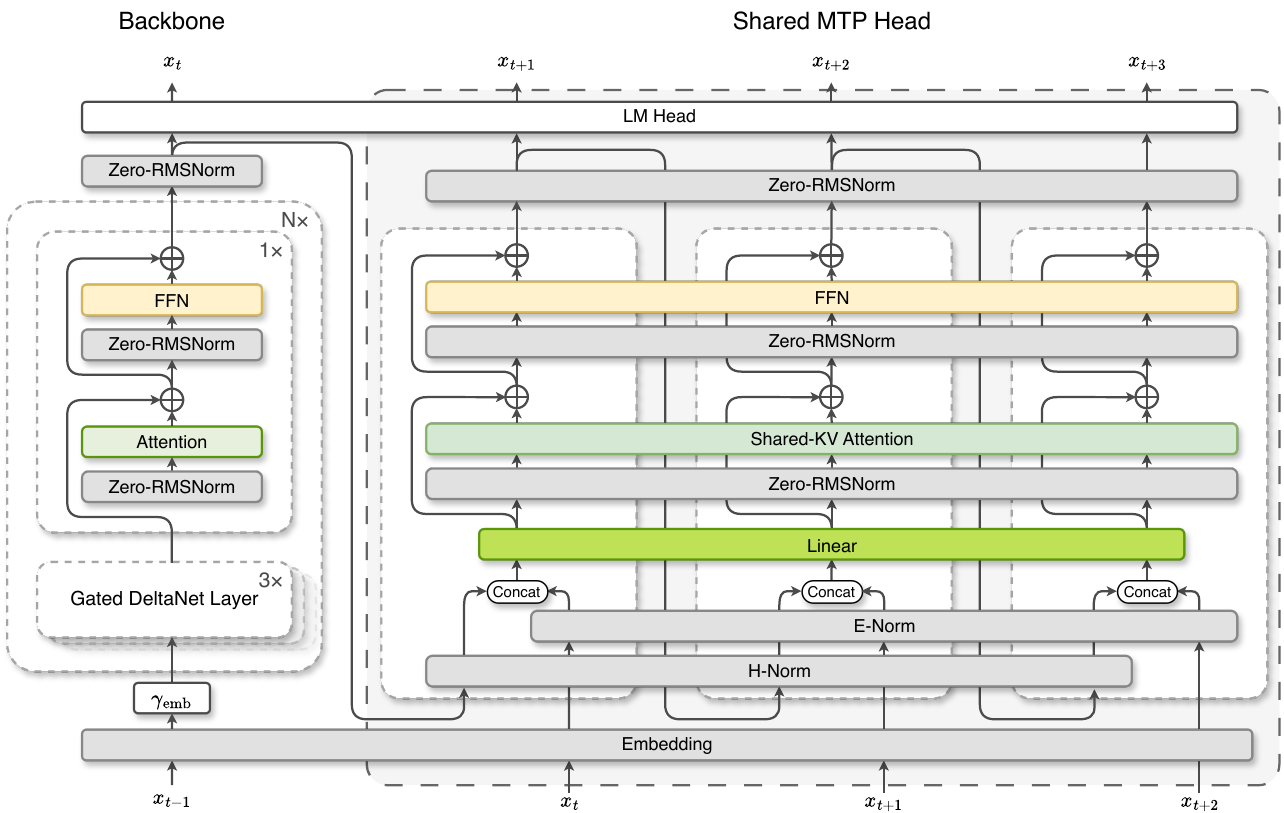}
    \caption{\textbf{The shared-KV MTP architecture.} All prediction depths reuse one MTP block, while keys and values are constructed only at the first depth and shared by subsequent depths.}
    \label{fig:glm_mtp_arch_0730}
\end{figure}

The shared-KV variant replaces the depth-specific modules $\{\mathcal{D}^{(1)},\ldots,\mathcal{D}^{(K)}\}$ with a single recurrently applied MTP block $\mathcal{D}_{\theta}$. The embedding normalization, hidden-state normalization, fusion projection, attention, feed-forward network, and output normalization are all shared across prediction depths:
\begin{equation}
    \boldsymbol{z}_t^{(k)}
    =
    \mathbf{W}_{\mathrm{fuse}}
    \left[
        \mathrm{Norm}_{e}\!\left(\mathbf{E}(x_{t+k})\right);
        \mathrm{Norm}_{h}\!\left(\boldsymbol{h}_t^{(k-1)}\right)
    \right].
\end{equation}
Parameter sharing alone reduces the auxiliary model size from approximately $K P_{\mathrm{MTP}}$ to $P_{\mathrm{MTP}}$, but naively reapplying the shared block would still reconstruct keys and values at every prediction depth. The shared-KV variant further removes this redundancy through \emph{first-step KV sharing}.

Specifically, after the input normalization inside $\mathcal{D}_{\theta}$, the first MTP depth constructs a static KV bank:
\begin{equation}
    \bar{\mathbf{K}}
    =
    \mathrm{RoPE}\!\left(
        \mathbf{W}_K\,\widetilde{\mathbf{Z}}^{(1)}
    \right),
    \qquad
    \bar{\mathbf{V}}
    =
    \mathbf{W}_V\,\widetilde{\mathbf{Z}}^{(1)},
    \label{eq:mtp_shared_kv}
\end{equation}
where $\widetilde{\mathbf{Z}}^{(1)}$ denotes the normalized first-depth fused states. Every depth retains its own query,
\begin{equation}
    \mathbf{Q}^{(k)}
    =
    \mathrm{RoPE}\!\left(
        \mathbf{W}_Q\,\widetilde{\mathbf{Z}}^{(k)}
    \right),
    \qquad
    \mathbf{A}^{(k)}
    =
    \mathrm{Attention}\!\left(
        \mathbf{Q}^{(k)}, \bar{\mathbf{K}}, \bar{\mathbf{V}}; \mathbf{M}_{\mathrm{causal}}
    \right),
    \label{eq:mtp_shared_attention}
\end{equation}
but depths $k>1$ neither execute $\mathbf{W}_K/\mathbf{W}_V$ nor append new entries to the KV bank. They only compute a new fusion state and query, attend to the same first-depth causal memory, and propagate the resulting hidden state to the next depth. Hence, shared-KV MTP preserves depth-dependent representations while sharing both parameters and memory.

For $K$ prediction depths, the conventional design performs $K$ sets of Q/K/V projections and maintains $K$ MTP caches. Shared-KV MTP performs $K$ query projections but only one set of K/V projections and maintains a single MTP cache. Ignoring the embedding and language-model head already shared with the backbone, the auxiliary parameter overhead is reduced by approximately a factor of $K$, while the K/V projection and cache costs are reduced from $K(C_Q+C_K+C_V)$ and $K M_{\mathrm{KV}}$ to $K C_Q+C_K+C_V$ and $M_{\mathrm{KV}}$, respectively. This reduction is particularly important on memory-constrained edge devices.

During training, the first depth builds an immutable full-sequence KV bank, which remains differentiable so that losses from deeper depths can update the first-depth fusion and K/V projections. During autoregressive inference, the backbone cache and the shared MTP cache are kept separate. Only the first MTP depth appends K/V for tokens that have been committed after verification; deeper depths access this cache in read-only mode. Consequently, rejected draft tokens never contaminate the persistent shared MTP cache, and no per-depth cache replay is required after a rejection.

The concrete parameter counts are summarized in Table~\ref{tab:mtp_params}. A single MTP block contains only 20.45M parameters, corresponding to approximately 3.1\% of the 653.70M-parameter main model, demonstrating the lightweight nature of the MTP module. With three prediction depths, standard MTP uses three such blocks, resulting in 715.05M total parameters, whereas shared-KV MTP reuses a single block and reduces the total parameter count to 674.15M.

\begin{table}[htbp]
    \centering
    \small
    \caption{\textbf{Parameter comparison between three-depth standard MTP and
    shared-KV MTP.} Component counts are approximate.}
    \label{tab:mtp_params}
    \begin{tabular}{lccc}
        \toprule
        \textbf{Architecture} & \textbf{Main Model} & \textbf{MTP Module} & \textbf{Total Parameters} \\
        \midrule
        Standard MTP ($K=3$) & 653.70M & 20.45M$\times$3 & 715.05M \\
        Shared-KV MTP ($K=3$) & 653.70M & 20.45M$\times$1 & 674.15M \\
        \bottomrule
    \end{tabular}
\end{table}

\subsubsection{Inference Speed}
\label{sec:xmtp_speed}
We measure the end-to-end decoding speedup of shared-KV X-MTP with $K=3$ in vLLM on a single RTX~4090 at batch size 1, the regime of on-device serving. The backbone verifies every draft exactly under greedy decoding, so X-MTP never changes the output distribution; the speedup is therefore a pure latency gain. Table~\ref{tab:xmtp_speed} reports decode throughput on 100 prompts from each of IFEval, GSM8K, MBPP, MATH-500, and ShareGPT, the last consisting of first user turns from real conversations. On mathematics and code, the drafter commits 3.3--3.6 tokens per backbone step and X-MTP accelerates decoding by $1.37$--$1.48\times$, lifting IronLLM-0.6B from 460 to 629--678 tokens/s. Open-ended conversation is harder to anticipate: on ShareGPT X-MTP commits 2.8 tokens per step for a $1.17\times$ speedup, and on IFEval, whose constrained instruction-following responses are least predictable from the backbone state, the gain drops to $1.04\times$. A single draft depth ($K=1$) does not pay for its drafting overhead ($0.83$--$0.93\times$), whereas each additional depth adds a further speedup, confirming that the recurrent shared-KV block remains accurate at deeper depths: at $K=3$ the per-depth acceptance rates on MATH-500 are 95\%, 89\%, and 82\%.

\begin{table}[ht]
    \centering
    \caption{\textbf{Decoding speedup of shared-KV X-MTP.} Batch size 1, greedy decoding with
    exact verification by the backbone, so drafts never change the output distribution.
    $N$ counts the prompts (of 100 sampled per benchmark) whose response ends with EOS in every
    run, $\tau$ is the mean number of tokens committed per backbone step at $K{=}3$, and TPS
    is the decode throughput with prefill excluded. RTX~4090, vLLM~0.17.1.}
    \label{tab:xmtp_speed}
    \small
    \begin{tabular}{l|cc|c|cc|ccc}
        \toprule
        \multirow{2}{*}{\textbf{Benchmark}} & \multirow{2}{*}{$\boldsymbol{N}$}
        & \multirow{2}{*}{\makecell{\textbf{Avg.}\\\textbf{out len}}} & \multirow{2}{*}{$\boldsymbol{\tau}$}
        & \multicolumn{2}{c|}{\textbf{Decode TPS}} & \multicolumn{3}{c}{\textbf{Speedup}} \\
        & & & & w/o MTP & $K=3$ & $K=1$ & $K=2$ & $K=3$ \\
        \midrule
        IFEval & 81 & 237 & 2.48 & 460 & 477 & 0.83$\times$ & 1.02$\times$ & \textbf{1.04$\times$} \\
        GSM8K & 99 & 139 & 3.42 & 460 & 662 & 0.92$\times$ & 1.27$\times$ & \textbf{1.44$\times$} \\
        MBPP & 99 & 93 & 3.27 & 460 & 629 & 0.91$\times$ & 1.23$\times$ & \textbf{1.37$\times$} \\
        MATH-500 & 93 & 317 & 3.55 & 460 & 678 & 0.93$\times$ & 1.29$\times$ & \textbf{1.48$\times$} \\
        ShareGPT & 87 & 268 & 2.82 & 460 & 540 & 0.87$\times$ & 1.11$\times$ & \textbf{1.17$\times$} \\
        \bottomrule
    \end{tabular}
\end{table}

\subsection{Lightweight Verification Head}

\subsubsection{Verification Head Design}
In deployment and inference speed tests with the MTP module, we observe that the wall-clock speedup falls noticeably short of the ideal gain implied by the acceptance rate, because predicting several tokens per step also inflates the latency of each step. Under the conventional drafting schedule, one decoding step must run all $K$ MTP depths to completion before any draft can be verified, and verification happens only in the next backbone forward pass. Each depth therefore pays for a full MTP block plus a vocabulary-sized output projection, the latter being by far the dominant cost. Since prefix acceptance discards every depth beyond the first rejection, this compute is spent unconditionally and becomes pure overhead whenever an early draft fails.

Verification itself is a second source of deployment complexity, as it requires speculatively advancing the backbone state over the proposed tokens and then rolling it back to the last accepted position. For softmax-attention layers a rollback reduces to truncating the KV cache, but IronLLM interleaves GDN layers whose fixed-size recurrent memory and short convolution state summarize the entire prefix in place and cannot be restored by slicing; an exact rollback demands either re-running the accepted prefix or snapshotting and restoring these states at every step. Both issues share a root cause: acceptance is decided too late and by too heavy a mechanism. We therefore attach a \emph{lightweight verification head} to each MTP depth, which predicts whether a draft token will be accepted at the moment that draft is produced, without consulting the backbone.

\begin{figure}[ht]
    \centering
    \includegraphics[width=0.9\textwidth]{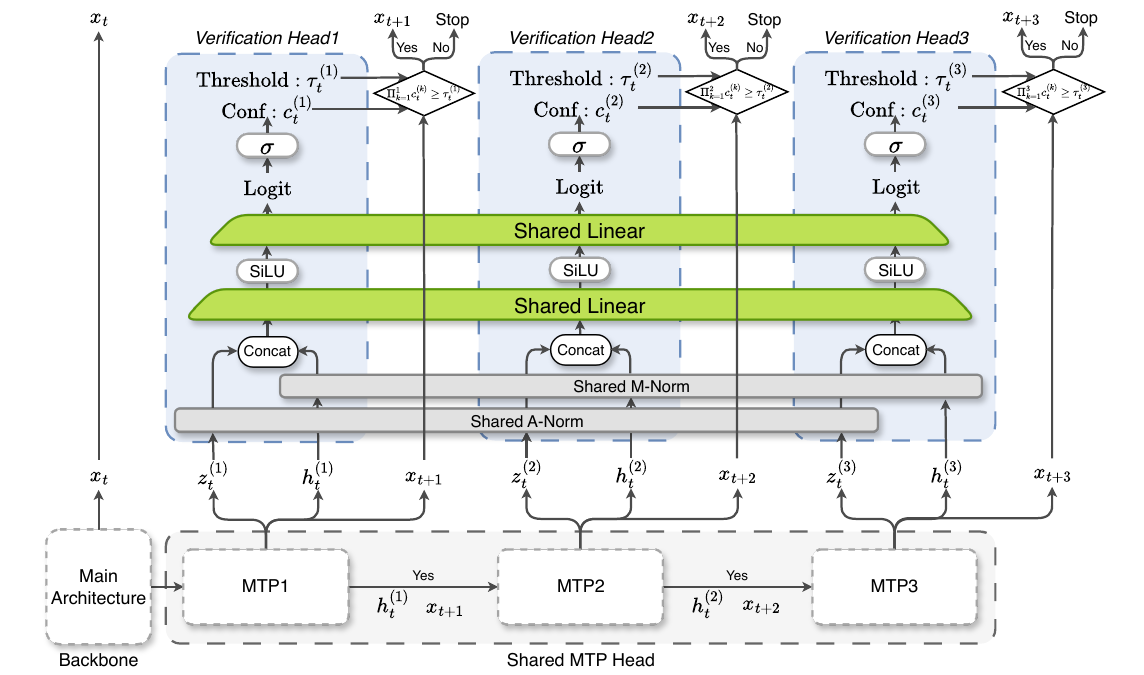}
    \caption{\textbf{The lightweight verification head.} At every MTP depth, a shared two-layer MLP consumes the aligned input hidden state and the output hidden state of that depth and emits a scalar acceptance confidence, which gates whether drafting proceeds to the next depth.}
    \label{fig:mtp_ver_head}
\end{figure}

\paragraph{Architecture.}
As illustrated in Figure~\ref{fig:mtp_ver_head}, the verification head is a two-layer MLP placed after each MTP depth. It observes the two hidden states that already characterize the prediction made at that depth: the aligned input state $\boldsymbol{z}_t^{(k)}$ produced by $\mathbf{W}_{\mathrm{fuse}}$, and the output state $\boldsymbol{h}_t^{(k)}$ from which the draft token is decoded. Following the same norm-then-concatenate fusion used by the MTP block, each stream is normalized separately before concatenation:
\begin{equation}
    c_t^{(k)}
    =
    \sigma\!\left(
        \mathbf{w}_{2}^{\top}
        \,\mathrm{SiLU}\!\left(
            \mathbf{W}_{1}
            \left[
                \mathrm{Norm}_{a}\!\left(\boldsymbol{z}_t^{(k)}\right);
                \mathrm{Norm}_{m}\!\left(\boldsymbol{h}_t^{(k)}\right)
            \right]
        \right)
    \right),
    \label{eq:ver_head}
\end{equation}

The scalar $c_t^{(k)}\in(0,1)$ estimates the probability that the draft proposed at depth $k$ matches the token the backbone would have committed. As with the shared-KV MTP block, one set of head parameters is reused across all $K$ depths, so the overhead is constant in $K$: 2.10M parameters, about 0.32\% of the 653.70M-parameter backbone and roughly one tenth of an MTP block.

\paragraph{Confidence-gated drafting.}
At inference time the head turns drafting from a fixed $K$-step pipeline into an adaptive one. Each depth is assigned its own threshold $\tau_k$, and once depth $k$ has produced its draft token the corresponding confidence is evaluated immediately, in parallel with decoding that token: clearing $\tau_k$ retains the draft and drafting continues to depth $k+1$, otherwise drafting terminates and depths $k+1,\ldots,K$ are never executed. To prevent errors from accumulating silently along the chain, the decision is made on the joint confidence, i.e.\ the product of the per-depth confidences accumulated so far:
\begin{equation}
    C_t^{(k)}
    =
    \prod_{j=1}^{k} c_t^{(j)},
    \qquad
    \hat{k}_t
    =
    \max\left\{
        k\in\{0,\ldots,K\}
        \;\middle|\;
        C_t^{(j)}\geq\tau_j\ \ \forall\, j\leq k
    \right\},
    \label{eq:joint_conf}
\end{equation}
with $\hat{k}_t$ the number of accepted drafts. Because each factor lies in $(0,1)$, $C_t^{(k)}$ is non-increasing in $k$ and approximates the probability that the entire draft prefix of length $k$ is correct, rather than that of one isolated depth. A locally plausible depth can thus no longer extend a prefix whose joint reliability has already decayed.

The scheme also removes the rollback problem entirely. Acceptance is settled before the backbone is invoked, so the backbone forward pass extends over the committed prefix only, and rejected drafts never reach the backbone KV cache, the GDN recurrent and convolution states, or the shared MTP cache. No truncation, replay, or state snapshotting is required at any point, which is precisely what makes speculative decoding practical for a hybrid linear-attention model on device.

\subsubsection{Training Strategy}
Following PIPO~\citep{tan2026pair}, we train the verification head in two stages. In both, the head is a purely auxiliary regressor: the two hidden states feeding Equation~\eqref{eq:ver_head} and the regression target are both detached, so its gradients never perturb the backbone or the MTP block. Adding the head is therefore behavior-preserving for the base model, and the overall objective simply extends Equation~\eqref{eq:mtp_loss} with one more weighted term,
\begin{equation}
    \mathcal{L}
    =
    \mathcal{L}_{\mathrm{AR}}
    +
    \lambda_{\mathrm{MTP}}\mathcal{L}_{\mathrm{MTP}}
    +
    \lambda_{\mathrm{ver}}\mathcal{L}_{\mathrm{ver}},
    \qquad
    \mathcal{L}_{\mathrm{ver}}
    =
    \frac{1}{K}
    \sum_{k=1}^{K}
    \frac{
        \sum_{t} m_t^{(k)}\,
        \mathrm{BCE}\!\left(c_t^{(k)}, y_t^{(k)}\right)
    }{
        \sum_{t} m_t^{(k)}
    },
    \label{eq:conf_loss}
\end{equation}
where $m_t^{(k)}$ is the same validity mask used by the MTP objective.

\paragraph{Stage I: probability matching during SFT.}
During SFT, MTP is trained with teacher forcing, so no on-policy draft tokens exist and the acceptance event is not observable. We instead supervise the head with the MTP head's own probability of the ground-truth continuation,
\begin{equation}
    y_t^{(k)}
    =
    \mathrm{sg}\!\left[\boldsymbol{p}_t^{(k)}(x_{t+k+1})\right],
    \label{eq:conf_target_sft}
\end{equation}
where $\mathrm{sg}[\cdot]$ denotes the stop-gradient operator. Under a deterministic verifier that commits the ground-truth token, this quantity is precisely the acceptance probability of the draft at depth $k$. It remains only a surrogate, since it is computed from teacher-forced inputs instead of self-generated drafts and scores the ground-truth token instead of the one the backbone would commit. Its value is that the signal is dense and essentially free at every valid position and depth, yielding a well-scaled head that initializes the sparser, draft-dependent objective of the next stage.

\paragraph{Stage II: acceptance matching during OPD.}
The on-policy distillation (OPD) stage removes both mismatches: drafts are generated by the MTP chain itself, and a teacher distribution acts as the verifier. As observed in PIPO, the OPD teacher plays exactly the role of the speculative-decoding verifier, under which a draft $\hat{y}$ proposed from $q$ and verified against $p$ is accepted with probability $\min\{1, p(\hat{y})/q(\hat{y})\}$. Taking the depth-$k$ MTP head as the proposal and the teacher as the verifier gives
\begin{equation}
    \hat{y}_t^{(k)} \sim \boldsymbol{p}_t^{(k)},
    \qquad
    y_t^{(k)}
    =
    \mathrm{sg}\!\left[
        \min\!\left\{
            1,\;
            \frac{
                \boldsymbol{p}_t^{\mathrm{tea}}\!\left(\hat{y}_t^{(k)}\right)
            }{
                \boldsymbol{p}_t^{(k)}\!\left(\hat{y}_t^{(k)}\right)
            }
        \right\}
    \right].
    \label{eq:ver_target_opd}
\end{equation}
The head stays detached here as well, so this objective leaves the distillation of the model itself untouched.

\subsubsection{Effectiveness of Verification Head}
\label{sec:ver_head_eval}
We jointly train the verification head during the MOPD stage. We evaluate the resulting head on three representative benchmarks—MMLU-Redux, IFEval, and GSM8K—with the inference threshold uniformly set to 0.9 across all depths. Table~\ref{tab:mtp_ver_head_eval} reports the evaluation results. We observe that for general QA tasks such as MMLU-Redux and IFEval, the verification head achieves nearly lossless performance while maintaining high acceptance rates. However, for complex reasoning tasks like GSM8K, using the verification head leads to noticeable score degradation. This is primarily because for tasks that require rigorous reasoning, such as mathematics and code generation, a single verification error introduced by the verification head can disrupt the model's reasoning chain, and the accumulated errors ultimately lead to incorrect final answers. More results—including the accuracy trajectory during joint training and a systematic sweep of the inference verification threshold are provided in Appendix~\ref{app:mtp_ver_head}.

\begin{table}[ht]
    \centering
    \caption{\textbf{Evaluation of the jointly trained verification head with the inference threshold uniformly set to 0.9.} Main Model denotes non-speculative decoding with the backbone; Avg.\ Accept.\ Len.\ is the average number of tokens committed per decoding step, and Match Rate is the fraction of confidence-accepted draft tokens that agree with the tokens committed by the backbone, both measured under confidence-gated MTP decoding.}
    \label{tab:mtp_ver_head_eval}
    \small
    \begin{tabular}{l|ccc|cc}
        \toprule
        \textbf{Benchmark} & \textbf{Main Model} & \textbf{w/ Ver.\ Head} & \textbf{$\Delta$} & \textbf{Avg.\ Accept.\ Len.} & \textbf{Match Rate} \\
        \midrule
        MMLU-Redux & 60.35 & 60.43 & $+$0.08 & 2.41 & 0.983 \\
        IFEval & 75.60 & 72.83 & $-$2.77 & 2.23 & 0.968 \\
        GSM8K & 78.62 & 66.26 & $-$12.36 & 2.12 & 0.967 \\
        \bottomrule
    \end{tabular}
\end{table}

\section{Conclusion, Limitation, and Future Work}

In this work, we present IronLLM-0.6B, a compact language model for efficient and capable on-device inference. It combines a hybrid attention architecture, a quality-oriented pre-training pipeline, and a post-training recipe for small models. Multi-Domain On-Policy Distillation integrates complementary capabilities from domain-specialized teachers without the negative transfer of parameter merging, and an Instruct-Only objective yields concise responses. The streamlined IronLLM-0.6B-Light improves deployment and quantization efficiency, while X-MTP delivers a $1.48\times$ decoding speedup via cross-step parameter and KV sharing. Together, these designs balance capability and efficiency, making the IronLLM models particularly well-suited to embodied systems with stringent compute, latency, and power constraints.

Several open challenges remain, pointing to clear directions for future work. Performance gaps persist on specialized mathematics and code-generation tasks, and the gains from our efficiency techniques may vary across hardware platforms, inference engines, and quantization settings. Future work will therefore pursue more capable and efficient edge-native architectures alongside data-efficient training recipes, extend the models toward compact multimodality and robot-side deployment, and broaden evaluation to production-grade edge hardware and embodied tasks, covering multilingual generation, safety, robustness, and tool use. We hope IronLLM models serve as a solid step toward bringing capable and efficient language models to embodied platforms.

\section*{Acknowledgments}

This work was made possible by the strong support of XPENG Robotics and the Foundation Model Team, along with the joint efforts of our research and engineering teams. We thank XPENG Robotics for providing computing and engineering resources, and for its continued commitment to building efficient foundation models for embodied intelligence. Special thanks go to all colleagues who contributed to data preparation, infrastructure support, model development, experimentation and evaluation.

\clearpage
\nocite{*}
\bibliographystyle{unsrtnat}
\bibliography{references}

@inproceedings{vaswani2017attention,
  title = {Attention Is All You Need},
  author = {Vaswani, Ashish and Shazeer, Noam and Parmar, Niki and Uszkoreit, Jakob and Jones, Llion and Gomez, Aidan N. and Kaiser, Lukasz and Polosukhin, Illia},
  booktitle = {Advances in Neural Information Processing Systems},
  year = {2017},
  url = {https://arxiv.org/abs/1706.03762}
}

@article{shao2024deepseekmath,
  title={Deepseekmath: Pushing the limits of mathematical reasoning in open language models},
  author={Shao, Zhihong and Wang, Peiyi and Zhu, Qihao and Xu, Runxin and Song, Junxiao and Bi, Xiao and Zhang, Haowei and Zhang, Mingchuan and Li, YK and Wu, Yang and others},
  journal={arXiv preprint arXiv:2402.03300},
  year={2024}
}

@article{deepseekai2024deepseekv3,
  title = {DeepSeek-V3 Technical Report},
  author = {{DeepSeek-AI}},
  journal = {arXiv preprint arXiv:2412.19437},
  year = {2024},
  url = {https://arxiv.org/abs/2412.19437}
}

@article{su2021roformer,
  title = {RoFormer: Enhanced Transformer with Rotary Position Embedding},
  author = {Su, Jianlin and Lu, Yu and Pan, Shengfeng and Murtadha, Ahmed and Wen, Bo and Liu, Yunfeng},
  journal = {arXiv preprint arXiv:2104.09864},
  year = {2021},
  url = {https://arxiv.org/abs/2104.09864}
}

@article{ainslie2023gqa,
  title = {GQA: Training Generalized Multi-Query Transformer Models from Multi-Head Checkpoints},
  author = {Ainslie, Joshua and Lee-Thorp, James and de Jong, Michiel and Zemlyanskiy, Yury and Lebron, Federico and Sanghai, Sumit},
  journal = {arXiv preprint arXiv:2305.13245},
  year = {2023},
  url = {https://arxiv.org/abs/2305.13245}
}

@inproceedings{hendrycks2021mmlu,
  title     = {Measuring Massive Multitask Language Understanding},
  author    = {Hendrycks, Dan and Burns, Collin and Basart, Steven
               and Zou, Andy and Mazeika, Mantas and Song, Dawn
               and Steinhardt, Jacob},
  booktitle = {International Conference on Learning Representations},
  year      = {2021}
}

@inproceedings{joshi2017triviaqa,
  title     = {{TriviaQA}: A Large Scale Distantly Supervised Challenge
               Dataset for Reading Comprehension},
  author    = {Joshi, Mandar and Choi, Eunsol and Weld, Daniel
               and Zettlemoyer, Luke},
  booktitle = {Proceedings of the 55th Annual Meeting of the
               Association for Computational Linguistics},
  pages     = {1601--1611},
  year      = {2017}
}

@article{clark2018arc,
  title   = {Think You Have Solved Question Answering?
             Try {ARC}, the {AI2} Reasoning Challenge},
  author  = {Clark, Peter and Cowhey, Isaac and Etzioni, Oren
             and Khot, Tushar and Sabharwal, Ashish
             and Schoenick, Carissa and Tafjord, Oyvind},
  journal = {arXiv preprint arXiv:1803.05457},
  year    = {2018}
}

@inproceedings{zellers2019hellaswag,
  title     = {{HellaSwag}: Can a Machine Really Finish Your Sentence?},
  author    = {Zellers, Rowan and Holtzman, Ari and Bisk, Yonatan
               and Farhadi, Ali and Choi, Yejin},
  booktitle = {Proceedings of the 57th Annual Meeting of the
               Association for Computational Linguistics},
  pages     = {4791--4800},
  year      = {2019}
}

@inproceedings{talmor2019commonsenseqa,
  title     = {{CommonsenseQA}: A Question Answering Challenge
               Targeting Commonsense Knowledge},
  author    = {Talmor, Alon and Herzig, Jonathan and Lourie, Nicholas
               and Berant, Jonathan},
  booktitle = {Proceedings of the 2019 Conference of the North American
               Chapter of the Association for Computational Linguistics:
               Human Language Technologies},
  pages     = {4149--4158},
  year      = {2019}
}

@inproceedings{mihaylov2018openbookqa,
  title     = {Can a Suit of Armor Conduct Electricity?
               A New Dataset for Open Book Question Answering},
  author    = {Mihaylov, Todor and Clark, Peter and Khot, Tushar
               and Sabharwal, Ashish},
  booktitle = {Proceedings of the 2018 Conference on Empirical Methods
               in Natural Language Processing},
  pages     = {2381--2391},
  year      = {2018}
}

@inproceedings{bisk2020piqa,
  title     = {{PIQA}: Reasoning about Physical Commonsense
               in Natural Language},
  author    = {Bisk, Yonatan and Zellers, Rowan and Le Bras, Ronan
               and Gao, Jianfeng and Choi, Yejin},
  booktitle = {Proceedings of the AAAI Conference on Artificial Intelligence},
  year      = {2020}
}

@inproceedings{sakaguchi2020winogrande,
  title     = {{WinoGrande}: An Adversarial Winograd Schema
               Challenge at Scale},
  author    = {Sakaguchi, Keisuke and Le Bras, Ronan
               and Bhagavatula, Chandra and Choi, Yejin},
  booktitle = {Proceedings of the AAAI Conference on Artificial Intelligence},
  year      = {2020}
}

@article{suzgun2022bbh,
  title   = {Challenging {BIG-Bench} Tasks and Whether
             Chain-of-Thought Can Solve Them},
  author  = {Suzgun, Mirac and Scales, Nathan and Sch{\"a}rli, Nathanael
             and Gehrmann, Sebastian and Tay, Yi and Chung, Hyung Won
             and Chowdhery, Aakanksha and Le, Quoc V.
             and Chi, Ed H. and Zhou, Denny and Wei, Jason},
  journal = {arXiv preprint arXiv:2210.09261},
  year    = {2022}
}

@article{li2023cmmlu,
  title   = {{CMMLU}: Measuring Massive Multitask Language
             Understanding in Chinese},
  author  = {Li, Haonan and Zhang, Yixuan and Koto, Fajri
             and Yang, Yifei and Zhao, Hai and Gong, Yeyun
             and Duan, Nan and Baldwin, Timothy},
  journal = {arXiv preprint arXiv:2306.09212},
  year    = {2023}
}

@article{huang2023ceval,
  title   = {{C-Eval}: A Multi-Level Multi-Discipline Chinese
             Evaluation Suite for Foundation Models},
  author  = {Huang, Yuzhen and Bai, Yuzhuo and Zhu, Zhihao
             and Zhang, Junlei and Zhang, Jinghan and Su, Tangjun
             and Liu, Junteng and Lv, Chuancheng and Zhang, Yikai
             and Lei, Jiayi and Fu, Yao and Sun, Maosong
             and He, Junxian},
  journal = {arXiv preprint arXiv:2305.08322},
  year    = {2023}
}

@article{sun2020c3,
  title   = {Investigating Prior Knowledge for Challenging
             Chinese Machine Reading Comprehension},
  author  = {Sun, Kai and Yu, Dian and Yu, Dong and Cardie, Claire},
  journal = {Transactions of the Association for Computational Linguistics},
  volume  = {8},
  pages   = {141--155},
  year    = {2020}
}

@article{schulman2017ppo,
  title = {Proximal Policy Optimization Algorithms},
  author = {Schulman, John and Wolski, Filip and Dhariwal, Prafulla and Radford, Alec and Klimov, Oleg},
  journal = {arXiv preprint arXiv:1707.06347},
  year = {2017},
  url = {https://arxiv.org/abs/1707.06347}
}

@article{zhou2023ifeval,
  title   = {Instruction-Following Evaluation for Large Language Models},
  author  = {Zhou, Jeffrey and Lu, Tianjian and Mishra, Swaroop
             and Brahma, Siddhartha and Basu, Sujoy and Luan, Yi
             and Zhou, Denny and Hou, Le},
  journal = {arXiv preprint arXiv:2311.07911},
  year    = {2023}
}

@article{cobbe2021gsm8k,
  title   = {Training Verifiers to Solve Math Word Problems},
  author  = {Cobbe, Karl and Kosaraju, Vineet and Bavarian, Mohammad
             and Chen, Mark and Jun, Heewoo and Kaiser, Lukasz
             and Plappert, Matthias and Tworek, Jerry
             and Hilton, Jacob and Nakano, Reiichiro
             and Hesse, Christopher and Schulman, John},
  journal = {arXiv preprint arXiv:2110.14168},
  year    = {2021}
}

@article{hendrycks2021math,
  title   = {Measuring Mathematical Problem Solving
             with the {MATH} Dataset},
  author  = {Hendrycks, Dan and Burns, Collin and Kadavath, Saurav
             and Arora, Akul and Basart, Steven and Tang, Eric
             and Song, Dawn and Steinhardt, Jacob},
  journal = {arXiv preprint arXiv:2103.03874},
  year    = {2021}
}

@article{chen2021humaneval,
  title   = {Evaluating Large Language Models Trained on Code},
  author  = {Chen, Mark and others},
  journal = {arXiv preprint arXiv:2107.03374},
  year    = {2021}
}

@article{austin2021mbpp,
  title   = {Program Synthesis with Large Language Models},
  author  = {Austin, Jacob and Odena, Augustus and Nye, Maxwell
             and Bosma, Maarten and Michalewski, Henryk
             and Dohan, David and Jiang, Ellen and Cai, Carrie
             and Terry, Michael and Le, Quoc V. and Sutton, Charles},
  journal = {arXiv preprint arXiv:2108.07732},
  year    = {2021}
}

@article{yang2025qwen3,
  title={Qwen3 technical report},
  author={Yang, An and Li, Anfeng and Yang, Baosong and Zhang, Beichen and Hui, Binyuan and Zheng, Bo and Yu, Bowen and Gao, Chang and Huang, Chengen and Lv, Chenxu and others},
  journal={arXiv preprint arXiv:2505.09388},
  year={2025}
}

@article{team2026kimi,
  title={Kimi K2. 5: Visual Agentic Intelligence},
  author={Team, Kimi and Bai, Tongtong and Bai, Yifan and Bao, Yiping and Cai, SH and Cao, Yuan and Charles, Y and Che, HS and Chen, Cheng and Chen, Guanduo and others},
  journal={arXiv preprint arXiv:2602.02276},
  year={2026}
}

@article{wang2025nemotron,
  title={Nemotron-cascade: Scaling cascaded reinforcement learning for general-purpose reasoning models},
  author={Wang, Boxin and Lee, Chankyu and Lee, Nayeon and Lin, Sheng-Chieh and Dai, Wenliang and Chen, Yang and Chen, Yangyi and Yang, Zhuolin and Liu, Zihan and Shoeybi, Mohammad and others},
  journal={arXiv preprint arXiv:2512.13607},
  year={2025}
}

@article{liu2025deepseek,
  title={Deepseek-v3. 2: Pushing the frontier of open large language models},
  author={Liu, Aixin and Mei, Aoxue and Lin, Bangcai and Xue, Bing and Wang, Bingxuan and Xu, Bingzheng and Wu, Bochao and Zhang, Bowei and Lin, Chaofan and Dong, Chen and others},
  journal={arXiv preprint arXiv:2512.02556},
  year={2025}
}

@article{zeng2025glm,
  title={Glm-4.5: Agentic, reasoning, and coding (arc) foundation models},
  author={Zeng, Aohan and Lv, Xin and Zheng, Qinkai and Hou, Zhenyu and Chen, Bin and Xie, Chengxing and Wang, Cunxiang and Yin, Da and Zeng, Hao and Zhang, Jiajie and others},
  journal={arXiv preprint arXiv:2508.06471},
  year={2025}
}

@article{wortsman2022model,
  title={Model soups: averaging weights of multiple fine-tuned models improves accuracy without increasing inference time},
  author={Wortsman, Mitchell and Ilharco, Gabriel and Gadre, Samir Yitzhak and Roelofs, Rebecca and Gontijo-Lopes, Raphael and Morcos, Ari S and Namkoong, Hongseok and Farhadi, Ali and Carmon, Yair and Kornblith, Simon and others},
  journal={arXiv preprint arXiv:2203.05482},
  year={2022}
}

@inproceedings{ilharco2022editing,
  title={Editing models with task arithmetic},
  author={Ilharco, Gabriel and Ribeiro, Marco Tulio and Wortsman, Mitchell and Schmidt, Ludwig and Hajishirzi, Hannaneh and Farhadi, Ali},
  booktitle={The Eleventh International Conference on Learning Representations},
  year={2022}
}

@inproceedings{agarwal2024policy,
  title={On-policy distillation of language models: Learning from self-generated mistakes},
  author={Agarwal, Rishabh and Vieillard, Nino and Zhou, Yongchao and Stanczyk, Piotr and Ramos Garea, Sabela and Geist, Matthieu and Bachem, Olivier},
  booktitle={International Conference on Learning Representations},
  volume={2024},
  pages={21246--21263},
  year={2024}
}

@inproceedings{gu2024minillm,
  title={Minillm: Knowledge distillation of large language models},
  author={Gu, Yuxian and Dong, Li and Wei, Furu and Huang, Minlie},
  booktitle={The twelfth international conference on learning representations},
  year={2024}
}

@article{ma2026mopd,
  title={Mopd: Multi-teacher on-policy distillation for capability integration in llm post-training},
  author={Ma, Wenhan and Wei, Jianyu and Zhao, Liang and Zhang, Hailin and Xiao, Bangjun and Li, Lei and Yang, Qibin and Gao, Bofei and Wang, Yudong and Li, Rang and others},
  journal={arXiv preprint arXiv:2606.30406},
  year={2026}
}

@article{zeng2026glm,
  title={Glm-5: from vibe coding to agentic engineering},
  author={Zeng, Aohan and Lv, Xin and Hou, Zhenyu and Du, Zhengxiao and Zheng, Qinkai and Chen, Bin and Yin, Da and Ge, Chendi and Huang, Chenghua and Xie, Chengxing and others},
  journal={arXiv preprint arXiv:2602.15763},
  year={2026}
}

@article{xiao2026mimo,
  title={Mimo-v2-flash technical report},
  author={Xiao, Bangjun and Xia, Bingquan and Yang, Bo and Gao, Bofei and Shen, Bowen and Zhang, Chen and He, Chenhong and Lou, Chiheng and Luo, Fuli and Wang, Gang and others},
  journal={arXiv preprint arXiv:2601.02780},
  year={2026}
}

@inproceedings{gu2025olmesstandardlanguagemodel,
  title={Olmes: A standard for language model evaluations},
  author={Gu, Yuling and Tafjord, Oyvind and Kuehl, Bailey and Haddad, Dany and Dodge, Jesse and Hajishirzi, Hannaneh},
  booktitle={Findings of the Association for Computational Linguistics: NAACL 2025},
  pages={5020--5048},
  year={2025}
}

@article{du2025understandingemergentabilitieslanguage,
  title={Understanding emergent abilities of language models from the loss perspective},
  author={Du, Zhengxiao and Zeng, Aohan and Dong, Yuxiao and Tang, Jie},
  journal={Advances in neural information processing systems},
  volume={37},
  pages={53138--53167},
  year={2024}
}

@article{li2025modelmergingpretraininglarge,
  title={Model merging in pre-training of large language models},
  author={Li, Yunshui and Ma, Yiyuan and Yan, Shen and Zhang, Chaoyi and Liu, Jing and Lu, Jianqiao and Xu, Ziwen and Chen, Mengzhao and Wang, Minrui and Zhan, Shiyi and others},
  journal={Advances in Neural Information Processing Systems},
  volume={38},
  pages={133668--133691},
  year={2026}
}

@article{wang2024mmlupro,
  title={Mmlu-pro: A more robust and challenging multi-task language understanding benchmark},
  author={Wang, Yubo and Ma, Xueguang and Zhang, Ge and Ni, Yuansheng and Chandra, Abhranil and Guo, Shiguang and Ren, Weiming and Arulraj, Aaran and He, Xuan and Jiang, Ziyan and others},
  journal={Advances in Neural Information Processing Systems},
  volume={37},
  pages={95266--95290},
  year={2024}
}

@inproceedings{gema2025mmluredux,
  title={Are we done with mmlu?},
  author={Gema, Aryo Pradipta and Leang, Joshua Ong Jun and Hong, Giwon and Devoto, Alessio and Mancino, Alberto Carlo Maria and Saxena, Rohit and He, Xuanli and Zhao, Yu and Du, Xiaotang and Madani, Mohammad Reza Ghasemi and others},
  booktitle={Proceedings of the 2025 Conference of the Nations of the Americas Chapter of the Association for Computational Linguistics: Human Language Technologies (Volume 1: Long Papers)},
  pages={5069--5096},
  year={2025}
}

@article{du2026supergpqa,
  title={Supergpqa: Scaling llm evaluation across 285 graduate disciplines},
  author={Du, Xeron and Yao, Yifan and Ma, Kaijing and Wang, Bingli and Zheng, Tianyu and Liu, Minghao and Liang, Yiming and Jin, Xiaolong and Wei, Zhenlin and Zheng, Chujie and others},
  journal={Advances in Neural Information Processing Systems},
  volume={38},
  year={2026}
}

@article{pyatkin2026ifbench,
  title={Generalizing verifiable instruction following},
  author={Pyatkin, Valentina and Malik, Saumya and Graf, Victoria and Ivison, Hamish and Huang, Shengyi and Dasigi, Pradeep and Lambert, Nathan and Hajishirzi, Hanna},
  journal={Advances in Neural Information Processing Systems},
  volume={38},
  year={2026}
}

@article{he2024multiif,
  title={Multi-if: Benchmarking llms on multi-turn and multilingual instructions following},
  author={He, Yun and Jin, Di and Wang, Chaoqi and Bi, Chloe and Mandyam, Karishma and Zhang, Hejia and Zhu, Chen and Li, Ning and Xu, Tengyu and Lv, Hongjiang and others},
  journal={arXiv preprint arXiv:2410.15553},
  year={2024}
}

@inproceedings{lightman2024math500,
  title={Let's verify step by step},
  author={Lightman, Hunter and Kosaraju, Vineet and Burda, Yuri and Edwards, Harrison and Baker, Bowen and Lee, Teddy and Leike, Jan and Schulman, John and Sutskever, Ilya and Cobbe, Karl},
  booktitle={International Conference on Learning Representations},
  volume={2024},
  pages={39578--39601},
  year={2024}
}

@article{dekoninck2026beyond,
  title={Beyond benchmarks: Matharena as an evaluation platform for mathematics with llms},
  author={Dekoninck, Jasper and Jovanovi{\'c}, Nikola and Gehrunger, Tim and R{\"o}gnvaldsson, K{\'a}ri and Petrov, Ivo and Sun, Chenhao and Vechev, Martin},
  journal={arXiv preprint arXiv:2605.00674},
  year={2026}
}

@inproceedings{jain2025livecodebench,
  title={Livecodebench: Holistic and contamination free evaluation of large language models for code},
  author={Jain, Naman and Gu, Alex and Li, Wen-Ding and Yan, Fanjia and Zhang, Tianjun and Wang, Sida and Solar-Lezama, Armando and Sen, Koushik and Stoica, Ion},
  booktitle={International Conference on Learning Representations},
  volume={2025},
  pages={58791--58831},
  year={2025}
}

@inproceedings{patil2025bfcl,
title={The Berkeley Function Calling Leaderboard (BFCL): From Tool Use to Agentic Evaluation of Large Language Models}, 
author={Patil, Shishir G. and Mao, Huanzhi and Cheng-Jie Ji, Charlie and Yan, Fanjia and Suresh, Vishnu and Stoica, Ion and E. Gonzalez, Joseph},
booktitle={Forty-second International Conference on Machine Learning},
year={2025},
}

@misc{evalscope_2024,
    title={{EvalScope}: Evaluation Framework for Large Models},
    author={ModelScope Team},
    year={2024},
    url={https://github.com/modelscope/evalscope}
}

@article{liquidai2025lfm2,
  title={LFM2 Technical Report},
  author={Liquid AI},
  journal={arXiv preprint arXiv:2511.23404},
  year={2025}
}

@article{minicpm4,
  title={Minicpm4: Ultra-efficient llms on end devices},
  author={MiniCPM, Team},
  journal={arXiv preprint arXiv:2506.07900},
  year={2025}
}

@misc{qwen3.5,
    title={{Qwen3.5}: Towards Native Multimodal Agents},
    author={{Qwen Team}},
    month={February},
    year={2026},
    url={https://qwen.ai/blog?id=qwen3.5}
}

@article{lin2025zebralogic,
  title={Zebralogic: On the scaling limits of llms for logical reasoning},
  author={Lin, Bill Yuchen and Bras, Ronan Le and Richardson, Kyle and Sabharwal, Ashish and Poovendran, Radha and Clark, Peter and Choi, Yejin},
  journal={arXiv preprint arXiv:2502.01100},
  year={2025}
}

@article{dubois2024alpacaevalv2,
  title={Length-controlled alpacaeval: A simple way to debias automatic evaluators},
  author={Dubois, Yann and Galambosi, Bal{\'a}zs and Liang, Percy and Hashimoto, Tatsunori B},
  journal={arXiv preprint arXiv:2404.04475},
  year={2024}
}

@article{li2024arenahard,
  title={From Crowdsourced Data to High-Quality Benchmarks: Arena-Hard and BenchBuilder Pipeline},
  author={Li, Tianle and Chiang, Wei-Lin and Frick, Evan and Dunlap, Lisa and Wu, Tianhao and Zhu, Banghua and Gonzalez, Joseph E and Stoica, Ion},
  journal={arXiv preprint arXiv:2406.11939},
  year={2024}
}

@article{kimiteam2025kimilinearexpressiveefficient,
  title={Kimi linear: An expressive, efficient attention architecture},
  author={Team, Kimi and Zhang, Yu and Lin, Zongyu and Yao, Xingcheng and Hu, Jiaxi and Meng, Fanqing and Liu, Chengyin and Men, Xin and Yang, Songlin and Li, Zhiyuan and others},
  journal={arXiv preprint arXiv:2510.26692},
  year={2025}
}

@inproceedings{yang2025gateddeltanetworksimproving,
  title={Gated delta networks: Improving mamba2 with delta rule},
  author={Yang, Songlin and Kautz, Jan and Hatamizadeh, Ali},
  booktitle={International Conference on Learning Representations},
  volume={2025},
  pages={29687--29707},
  year={2025}
}

@article{merrill2026olmohybridtheorypractice,
  title={Olmo hybrid: From theory to practice and back},
  author={Merrill, William and Li, Yanhong and Romero, Tyler and Svete, Anej and Costello, Caia and Dasigi, Pradeep and Groeneveld, Dirk and Heineman, David and Kuehl, Bailey and Lambert, Nathan and others},
  journal={arXiv preprint arXiv:2604.03444},
  year={2026}
}

@article{kimiteam2026kimik3openfrontier,
  title={Kimi k3: Open frontier intelligence},
  author={Team, Kimi and Bai, Tongtong and Bai, Yifan and Bao, Yiping and Cai, Jianfeng and Cai, Xinyuan and Cao, Peizhou and Cao, Yuxuan and Chai, Ziwei and Charles, Y and others},
  journal={arXiv preprint arXiv:2607.24653},
  year={2026}
}

@inproceedings{dehghani2023scalingvisiontransformers22,
  title={Scaling vision transformers to 22 billion parameters},
  author={Dehghani, Mostafa and Djolonga, Josip and Mustafa, Basil and Padlewski, Piotr and Heek, Jonathan and Gilmer, Justin and Steiner, Andreas Peter and Caron, Mathilde and Geirhos, Robert and Alabdulmohsin, Ibrahim and others},
  booktitle={International conference on machine learning},
  pages={7480--7512},
  year={2023},
  organization={PMLR}
}

@article{jiang2023prermsnormprecrmsnormtransformersequivalent,
  title={Pre-rmsnorm and pre-crmsnorm transformers: equivalent and efficient pre-ln transformers},
  author={Jiang, Zixuan and Gu, Jiaqi and Zhu, Hanqing and Pan, David},
  journal={Advances in Neural Information Processing Systems},
  volume={36},
  pages={45777--45793},
  year={2023}
}

@inproceedings{zhu2025transformersnormalization,
  title={Transformers without normalization},
  author={Zhu, Jiachen and Chen, Xinlei and He, Kaiming and LeCun, Yann and Liu, Zhuang},
  booktitle={2025 IEEE/CVF Conference on Computer Vision and Pattern Recognition (CVPR)},
  pages={14901--14911},
  year={2025},
  organization={IEEE}
}

@article{hu2024minicpmunveilingpotentialsmall,
  title={Minicpm: Unveiling the potential of small language models with scalable training strategies},
  author={Hu, Shengding and Tu, Yuge and Han, Xu and He, Chaoqun and Cui, Ganqu and Long, Xiang and Zheng, Zhi and Fang, Yewei and Huang, Yuxiang and Zhao, Weilin and others},
  journal={arXiv preprint arXiv:2404.06395},
  year={2024}
}

@article{loshchilov2019decoupledweightdecayregularization,
  title={Decoupled weight decay regularization},
  author={Loshchilov, Ilya and Hutter, Frank},
  journal={arXiv preprint arXiv:1711.05101},
  year={2017}
}

@article{ding2024fewertruncationsimprovelanguage,
  title={Fewer truncations improve language modeling},
  author={Ding, Hantian and Wang, Zijian and Paolini, Giovanni and Kumar, Varun and Deoras, Anoop and Roth, Dan and Soatto, Stefano},
  journal={arXiv preprint arXiv:2404.10830},
  year={2024}
}

@inproceedings{dao2023flashattention2fasterattentionbetter,
  title={Flashattention-2: Faster attention with better parallelism and work partitioning},
  author={Dao, Tri},
  booktitle={International Conference on Learning Representations},
  volume={2024},
  pages={35549--35562},
  year={2024}
}

@misc{yang2024fla,
  title  = {FLA: A Triton-Based Library for Hardware-Efficient Implementations of Linear Attention Mechanism},
  author = {Yang, Songlin and Zhang, Yu},
  url    = {https://github.com/fla-org/flash-linear-attention},
  month  = jan,
  year   = {2024}
}

@inproceedings{hsu2025ligerkernel,
title={Liger-Kernel: Efficient Triton Kernels for {LLM} Training},
author={Pin-Lun Hsu and Yun Dai and Vignesh Kothapalli and Qingquan Song and Shao Tang and Siyu Zhu and Steven Shimizu and Shivam Sahni and Haowen Ning and Yanning Chen and Zhipeng Wang},
booktitle={Championing Open-source DEvelopment in ML Workshop @ ICML25},
year={2025},
url={https://openreview.net/forum?id=36SjAIT42G}
}

@article{qin2024lightningattention2,
  title={Lightning Attention-2: A Free Lunch for Handling Unlimited Sequence Lengths in Large Language Models},
  author={Qin, Zhen and Sun, Weigao and Li, Dong and Shen, Xuyang and Sun, Weixuan and Zhong, Yiran},
  journal={arXiv preprint arXiv:2401.04658},
  year={2024},
  url={https://arxiv.org/abs/2401.04658}
}

@inproceedings{press2022alibi,
  title={Train Short, Test Long: Attention with Linear Biases Enables Input Length Extrapolation},
  author={Press, Ofir and Smith, Noah A. and Lewis, Mike},
  booktitle={International Conference on Learning Representations},
  year={2022},
  url={https://arxiv.org/abs/2108.12409}
}

@misc{qiu2026unifiedviewattentionresidual,
      title={A Unified View of Attention and Residual Sinks: Outlier-Driven Rescaling is Essential for Transformer Training}, 
      author={Zihan Qiu and Zeyu Huang and Kaiyue Wen and Peng Jin and Bo Zheng and Yuxin Zhou and Haofeng Huang and Zekun Wang and Xiao Li and Huaqing Zhang and Yang Xu and Haoran Lian and Siqi Zhang and Rui Men and Jianwei Zhang and Ivan Titov and Dayiheng Liu and Jingren Zhou and Junyang Lin},
      year={2026},
      eprint={2601.22966},
      archivePrefix={arXiv},
      primaryClass={cs.CL},
      url={https://arxiv.org/abs/2601.22966}, 
}

@misc{ramachandran2017searchingactivationfunctions,
      title={Searching for Activation Functions}, 
      author={Prajit Ramachandran and Barret Zoph and Quoc V. Le},
      year={2017},
      eprint={1710.05941},
      archivePrefix={arXiv},
      primaryClass={cs.NE},
      url={https://arxiv.org/abs/1710.05941}, 
}

@misc{choi2018pactparameterizedclippingactivation,
      title={PACT: Parameterized Clipping Activation for Quantized Neural Networks}, 
      author={Jungwook Choi and Zhuo Wang and Swagath Venkataramani and Pierce I-Jen Chuang and Vijayalakshmi Srinivasan and Kailash Gopalakrishnan},
      year={2018},
      eprint={1805.06085},
      archivePrefix={arXiv},
      primaryClass={cs.CV},
      url={https://arxiv.org/abs/1805.06085}, 
}

@misc{cao2026opencompassuniversalevaluationplatform,
      title={OpenCompass: A Universal Evaluation Platform for Large Language Models}, 
      author={Maosong Cao and Kai Chen and Haodong Duan and others},
      year={2026},
      eprint={2605.19276},
      archivePrefix={arXiv},
      primaryClass={cs.CL},
      url={https://arxiv.org/abs/2605.19276}, 
}

@article{tan2026pair,
  title={Pair-In, Pair-Out: Latent Multi-Token Prediction for Efficient LLMs},
  author={Tan, Wenhui and Li, Minghao and Ma, Xiaoqian and Fan, Siqi and Huang, Xiusheng and Zhang, Liujie and Song, Ruihua and Chen, Weihang},
  journal={arXiv preprint arXiv:2605.27255},
  year={2026}
}

@misc{olmo2026olmo3,
      title={Olmo 3}, 
      author={Team Olmo and Allyson Ettinger and Amanda Bertsch and others},
      year={2026},
      eprint={2512.13961},
      archivePrefix={arXiv},
      primaryClass={cs.CL},
      url={https://arxiv.org/abs/2512.13961}, 
}

\clearpage
\appendix
\section{Data Contamination Details}
\label{sec:data_contam}
This appendix provides additional details of our benchmark contamination analysis, together with the corresponding experimental results.

\paragraph{N-gram Overlap Detection.}
We detect potential benchmark contamination between the pre-training corpus and evaluation benchmarks using an IDF-weighted 5-gram overlap framework. An inverted 5-gram index is constructed over benchmark question stems to efficiently retrieve candidate training documents with high lexical overlap.

For each retrieved candidate, both question and answer overlaps are jointly considered. We define the contamination score as
\begin{equation}
\mathrm{sc}=0.75\times\mathrm{stem\_idf\_overlap}
+0.25\times\mathrm{aio},
\end{equation}
where $\mathrm{stem\_idf\_overlap}$ denotes the IDF-weighted lexical overlap between question stems, and $\mathrm{aio}$ (answer IDF overlap) measures lexical consistency between reference answers.

Candidate pairs with retrieval scores above 0.3 are retained for further verification. Samples are considered contaminated if $\mathrm{sc}\ge0.8$ or $\mathrm{aio}\ge0.5$, while borderline cases are resolved through benchmark-specific rules and manual inspection.

\paragraph{Embedding-based Semantic Retrieval.}
To identify benchmark contamination beyond lexical overlap, we further perform embedding-based semantic retrieval on the SFT corpus. Each SFT instruction is compared with benchmark questions using cosine similarity between sentence embeddings, allowing semantically similar but lexically different samples to be retrieved.

Candidate pairs with cosine similarity above 0.8 are further verified using an LLM as a judge, followed by human inspection. A sample is considered contaminated only when both the underlying semantics and evaluation target are consistent with the benchmark item, while semantically related but independently constructed examples are excluded from contamination statistics.

\paragraph{Results}
Table~\ref{tab:contamination} reports IronLLM-0.6B performance on the original and non-contaminated subsets across the evaluated benchmarks. Overall, the differences are small on most benchmarks, with no systematic performance degradation after decontamination.

\begin{table}[ht]
    \centering
    \caption{\textbf{Contamination Analysis.}
    Contaminated samples are identified as the union of N-gram
    overlap detection against the pretraining corpus and embedding-based
    semantic retrieval against the SFT data.}
    \label{tab:contamination}
    \small
    \begin{tabular}{lccc}
        \toprule
        \multirow[c]{2}{*}[-0.5ex]{\textbf{Test Set}}
        & \multicolumn{3}{c}{\textbf{IronLLM-0.6B}} \\
        \cmidrule(lr){2-4}
        & \textbf{Orig.} & \textbf{Non-Contam.} & $\boldsymbol{\Delta}$ \\
        \midrule
MMLU-Pro      & 42.1  & 43.1  & $+1.0$ \\
MMLU-Redux    & 60.4  & 61.0  & $+0.6$ \\
C-Eval        & 47.3  & 47.0  & $-0.3$ \\
CMMLU         & 49.4  & 49.7  & $+0.3$ \\
IFEval        & 75.6  & 75.7  & $+0.1$ \\
GSM8K         & 78.6  & 76.9  & $-1.7$ \\
MBPP          & 45.0  & 47.3  & $+2.3$ \\
BBH           & 37.6  & 37.0  & $-0.6$ \\
        \bottomrule
    \end{tabular}
\end{table}

\section{Training and Evaluation Details}

\subsection{Pre-training Loss Curve}

\begin{figure}[htbp]
    \centering
    \includegraphics[width=1.0\textwidth]{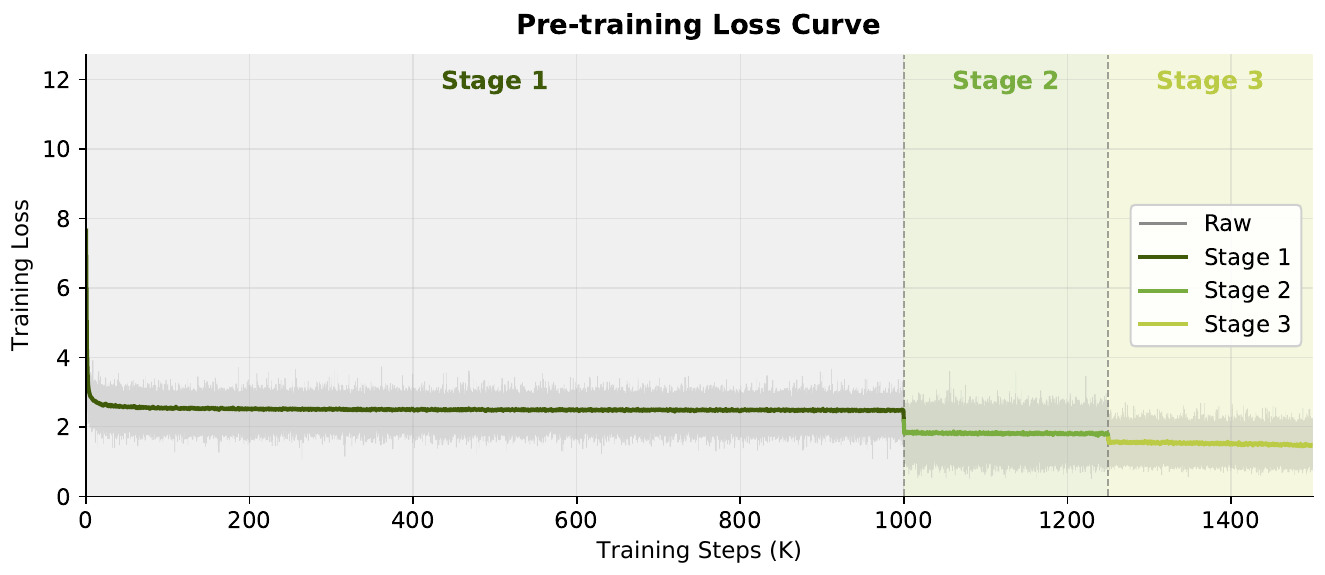}
    \caption{\textbf{Pre-training loss curve across the three training stages.}
    The gray trace shows the raw per-step loss and the colored curves show the
    loss smoothed with a rolling average over 1K steps. The loss decreases
    smoothly throughout training.}
    \label{fig:appendix_full_training_loss}
\end{figure}

Figure~\ref{fig:appendix_full_training_loss} shows the training loss over the
full pre-training run (1.5M steps, 6.2 trillion tokens in total). Training remains
stable throughout. After warmup, the loss decreases smoothly with no loss spikes or divergence observed.
The downward jumps at stage boundaries reflect shifts in data composition toward higher proportions of math, code, and reasoning data. During the decay phase of Stage~3, the loss further decreases corresponds to the learning-rate decay phase of the WSD schedule.

\subsection{Base Model Evaluation Protocol}
\label{app:base_evaluation}
Base model evaluation can be divided into two categories: perplexity-based metrics and free-generation-based metrics. Perplexity-based approaches further split into the multiple-choice formulation (MCF), which requires the model to choose among explicit options (e.g., A/B/C/D), and the cloze formulation (CF), which compares likelihoods of candidate completions without exposing options in the prompt.

Prior work~\cite{gu2025olmesstandardlanguagemodel, du2025understandingemergentabilitieslanguage} shows that MCF is especially difficult in early training, particularly for small models. We therefore use CF with benchmark-specific probability normalization for early-stage diagnostics in from-scratch runs, and report final base-model results using MCF. Unlike prior work that selectively reports the more favorable metric for each benchmark, we keep this protocol fixed throughout. We view MCF as better suited to practical option-based use cases once the model is mature. For the free-generation benchmarks, we apply minor prompt refinement and more carefully selected few-shot examples to reduce benchmark-specific sensitivity.

\begin{figure}[htbp]
    \centering
    \includegraphics[width=1.0\textwidth]{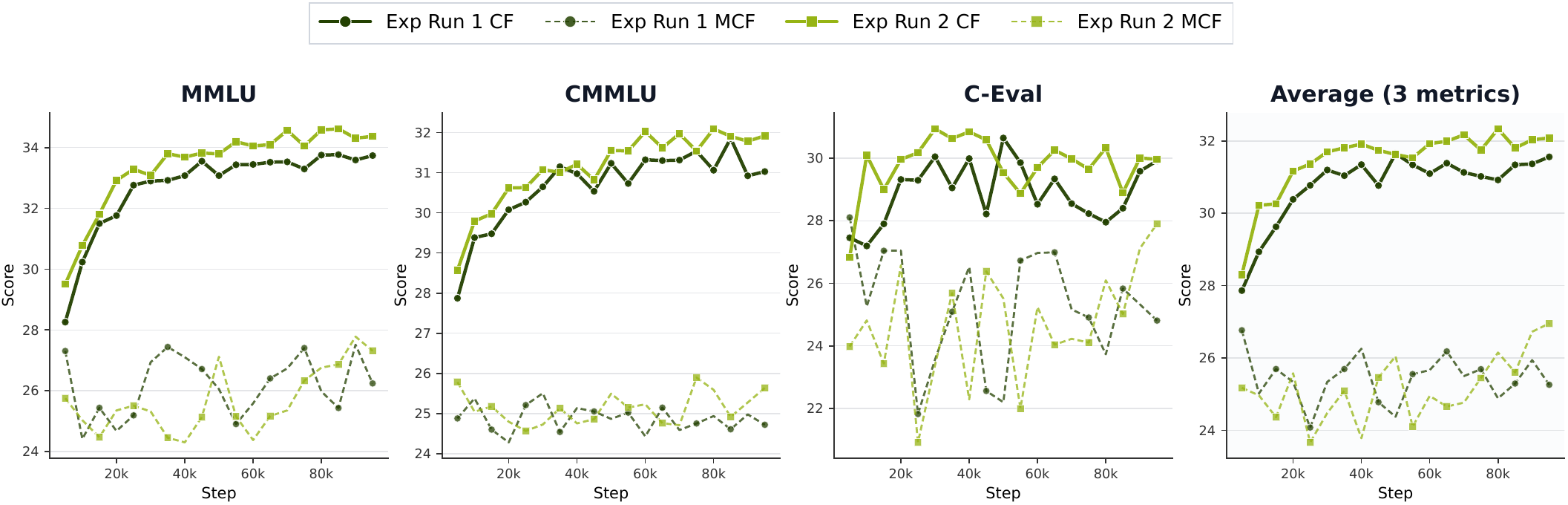}
    \caption{\textbf{Comparison of CF and MCF evaluation trajectories} for two pretraining runs conducted from scratch.}
    \label{fig:appendix_pretrain_evaluation_1}
\end{figure}

Figure~\ref{fig:appendix_pretrain_evaluation_1} illustrates the distinction on MMLU, CMMLU, C-Eval, and their average. CF curves from two independent runs are smooth and clearly separable, while MCF curves are highly variable, frequently cross, and remain near the 25\% random baseline, indicating that the model struggles with four-way multiple-choice questions during early training.

\subsection{MOPD Training Curves}
\label{app:mopd_curves}
\begin{figure}[htbp]
    \vspace{-1em}
    \centering
    \includegraphics[width=0.49\textwidth]{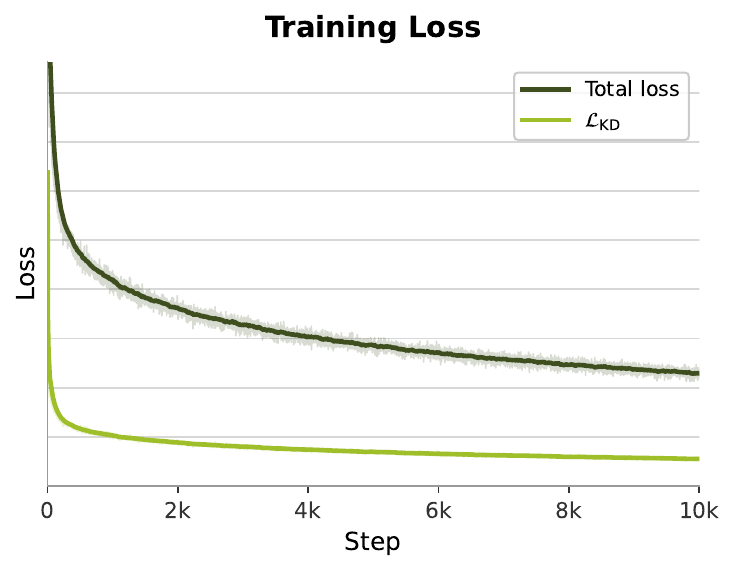}
    \hfill
    \includegraphics[width=0.49\textwidth]{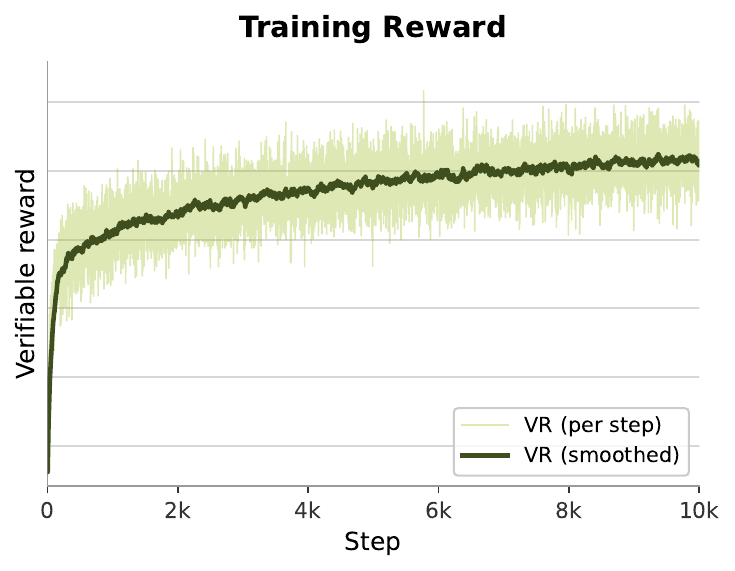}
    \vspace{-1em}
    \caption{\textbf{MOPD training curves over the first 10K steps.} Left: total
    optimized loss (dark) and the distillation term $\mathcal{L}_{\mathrm{KD}}$
    alone (light); the warm-up steps, where the total loss peaks at $6.4$, fall
    outside the plotted $y$-range. Right: batch-mean verifiable reward.
    Faint traces show raw per-step values, bold curves an exponential moving
    average (decay $0.97$).}
    \label{fig:appendix_mopd_curves}
    \vspace{-1em}
\end{figure}

Figure~\ref{fig:appendix_mopd_curves} shows the loss and reward over the first 10K
steps of MOPD training. The distillation term decreases from $1.59$ to $0.14$,
i.e.\ the student's average per-token disagreement with its routed teacher is
reduced by roughly an order of magnitude, and the total loss follows the same
trajectory, falling from $6.41$ to $0.57$. Meanwhile the verifiable reward rises
from near zero to $0.41$, steeply over the first few hundred steps and then slowly
but steadily. Training is stable throughout, with no loss spikes or reward
collapse, and the two objectives improve jointly rather than trading off against
each other.

\subsection{Additional Results of the Verification Head}
\label{app:mtp_ver_head}
This appendix supplements the verification head evaluation in Section~\ref{sec:ver_head_eval} with two additional studies of the jointly trained verification head: (i) the accuracy trajectory observed during the MOPD stage, and (ii) a systematic sweep of the inference confidence threshold, covering accuracy, average acceptance length, and confidence--verification consistency.

\begin{figure}[ht]
    \centering
    \includegraphics[width=0.86\textwidth]{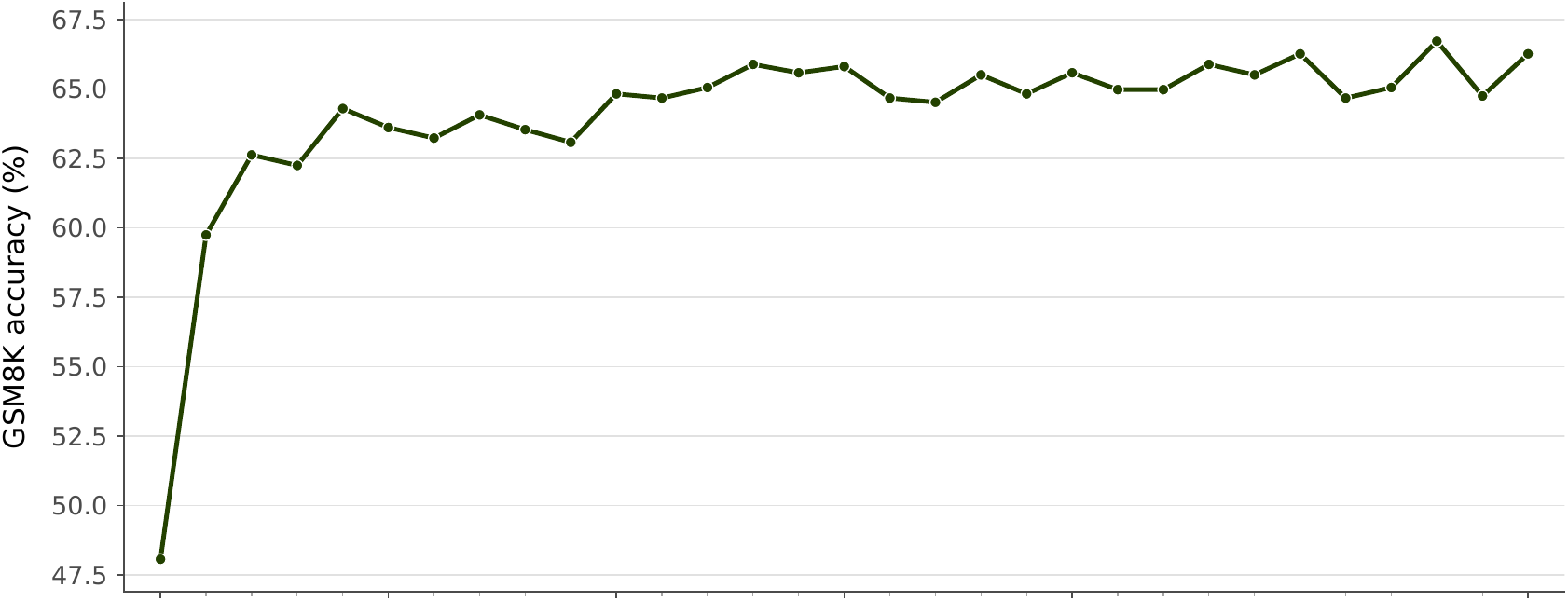}
    \vspace{-0.5em}
    \caption{\textbf{GSM8K accuracy of MTP inference w/ verification head (uniform threshold $\tau=0.9$) during MOPD training.} The accuracy rises steeply in the early training phase and saturates into a stable plateau by the end of the stage.}
    \label{fig:mtp_ver_gsm8k_training}
\end{figure}

\paragraph{Accuracy during Joint Training.}
During the MOPD stage, we evaluate confidence-gated MTP decoding on GSM8K at every epoch with the uniform threshold $\tau=0.9$ used in Section~\ref{sec:ver_head_eval}. Figure~\ref{fig:mtp_ver_gsm8k_training} plots the resulting trajectory. Starting from a low level at MOPD initialization, the accuracy rises steeply within the first epoch of on-policy distillation, keeps climbing over the next few epochs, and then saturates into a stable plateau for the remainder of the stage. Most of the total gain is realized within the first epoch, indicating that the on-policy objective rapidly aligns the MTP chain, and hence the acceptance events that supervise the head, with the teacher distribution.

\begin{figure}[ht]
    \centering
    \includegraphics[width=\textwidth]{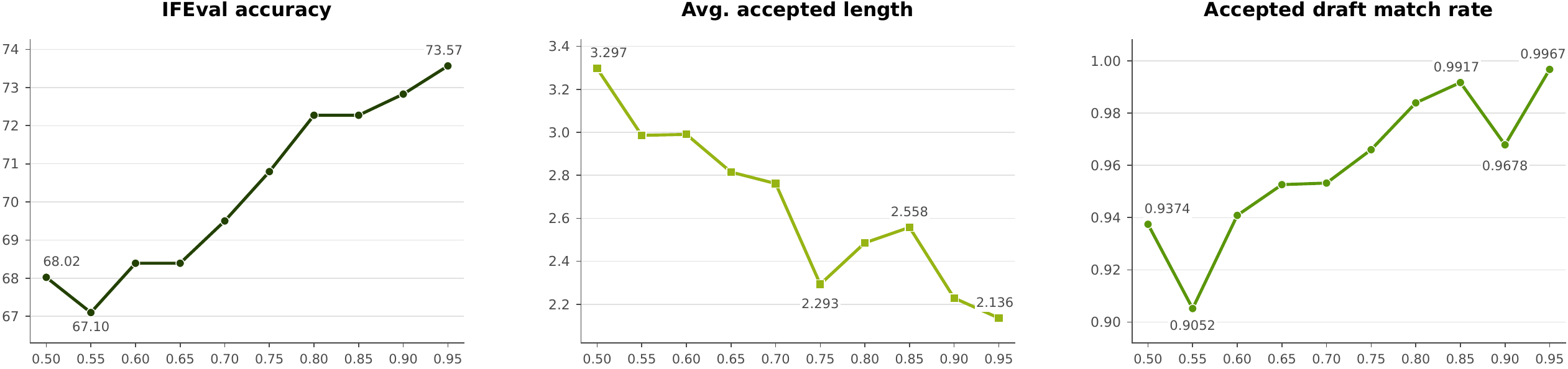}
    \vspace{-1em}
    \caption{\textbf{Effect of the uniform confidence threshold $\tau$ on IFEval:} accuracy (left), average acceptance length (middle), and confidence--verification match rate (right). Ten uniform thresholds from 0.50 to 0.95 are evaluated.}
    \label{fig:mtp_ver_ifeval_thr_sweep}
\end{figure}

\paragraph{Effect of the Confidence Threshold.}
The inference threshold controls how aggressively drafts are admitted, and hence the trade-off between the number of tokens committed per step and the risk of accepting an incorrect draft. Using the final MOPD checkpoint, we sweep the uniform per-depth threshold $\tau_k \equiv \tau$ from 0.50 to 0.95 on IFEval. Figure~\ref{fig:mtp_ver_ifeval_thr_sweep} plots the accuracy, the average acceptance length, and the confidence--verification match rate for the ten uniform settings.

The sweep exhibits a clear accuracy--speed trade-off. Relaxing the threshold admits more draft tokens per step but costs accuracy, whereas stricter settings progressively close the gap to the non-speculative baseline while still committing multiple tokens per step. The match rate improves accordingly as the threshold tightens, confirming that the head's confidence scores are informative: drafts admitted with high confidence are almost always verified as correct, and at the strict end of the sweep nearly every accepted draft agrees with the backbone.

\end{document}